\UseRawInputEncoding
\documentclass[journal]{IEEEtran}
\usepackage{graphicx} 
\usepackage{subfigure}
\usepackage{amsmath,amssymb,amsfonts}
\usepackage{amsthm,amsmath,amssymb}
\usepackage{mathrsfs}
\usepackage{bm}
\usepackage{algorithmic}
\usepackage{booktabs}
\usepackage[switch]{lineno}
\usepackage{float}
\usepackage{url}
\usepackage{epstopdf}
\newtheorem{thm}{Theorem}[section]
\newtheorem{lem}[thm]{Lemma}

\usepackage{color}

\ifCLASSINFOpdf
\else
\fi
\begin{document}
%



\title{EgPDE-Net: Building Continuous Neural Networks for Time Series Prediction with Exogenous Variables}

%
%
%
\author{Penglei~Gao,
        Xi~Yang,
        Rui~Zhang,
        Ping Guo,
        John Y. Goulermas,
        and~Kaizhu Huang*
\thanks{Penglei Gao and John Y. Goulermas are with the Department of Computer Science, University of Liverpool, Liverpool L69 7ZX, U.K. (Email: P.Gao6@liverpool.ac.uk;J.Y.Goulermas@liverpool.ac.uk)}
\thanks{Xi Yang is with the Department of Intelligent Science, Xi'an Jiaotong-Liverpool University, Suzhou 215123, China. (Email: xi.yang01@xjtlu.edu.cn)}
\thanks{Rui Zhang is with the Department of Foundational Mathematics, Xi'an Jiaotong-Liverpool University, Suzhou 215123, China. (Email: rui.zhang02@xjtlu.edu.cn)}
\thanks{Ping Guo is with the Department of school of systems science, Beijing Normal University, Beijing 100875, China. (Email: pguo@bnu.edu.cn)}
\thanks{*Corresponding Author: Kaizhu Huang is with the Institute of Applied Physical Sciences and Engineering, Duke Kunshan University, No. 8 Duke Avenue, Kunshan 215316. (Email: kaizhu.huang@dukekunshan.edu.cn)}
}

%



\maketitle

\begin{abstract}
While exogenous variables have a major impact on performance improvement in time series analysis, inter-series correlation and time dependence among them are rarely considered in the present continuous methods. The dynamical systems of multivariate time series could be modelled with complex unknown partial differential equations (PDEs) which play a prominent role in many disciplines of science and engineering.
In this paper, we propose a continuous-time model for arbitrary-step prediction to learn an unknown PDE system in multivariate time series whose governing equations are parameterised by self-attention and gated recurrent neural networks. The proposed model, \underline{E}xogenous-\underline{g}uided \underline{P}artial \underline{D}ifferential \underline{E}quation Network (EgPDE-Net), takes account of the relationships among the exogenous variables and their effects on the target series. Importantly, the model can be reduced into a regularised ordinary differential equation (ODE) problem with special designed regularisation guidance, which makes the PDE problem tractable to obtain numerical solutions and feasible to predict multiple future values of the target series at arbitrary time points.
Extensive experiments demonstrate that our proposed model could achieve competitive accuracy over strong baselines: on average, it outperforms the best baseline by reducing $9.85\%$ on RMSE and $13.98\%$ on MAE for arbitrary-step prediction.
\end{abstract}


%
\IEEEpeerreviewmaketitle

\section{Introduction}
\begin{figure}[!htbp]
	\centering
	\subfigure[Multi-step prediction by discrete neural networks.]{
		\label{diff.sub2}
		\includegraphics[width=0.4\textwidth]{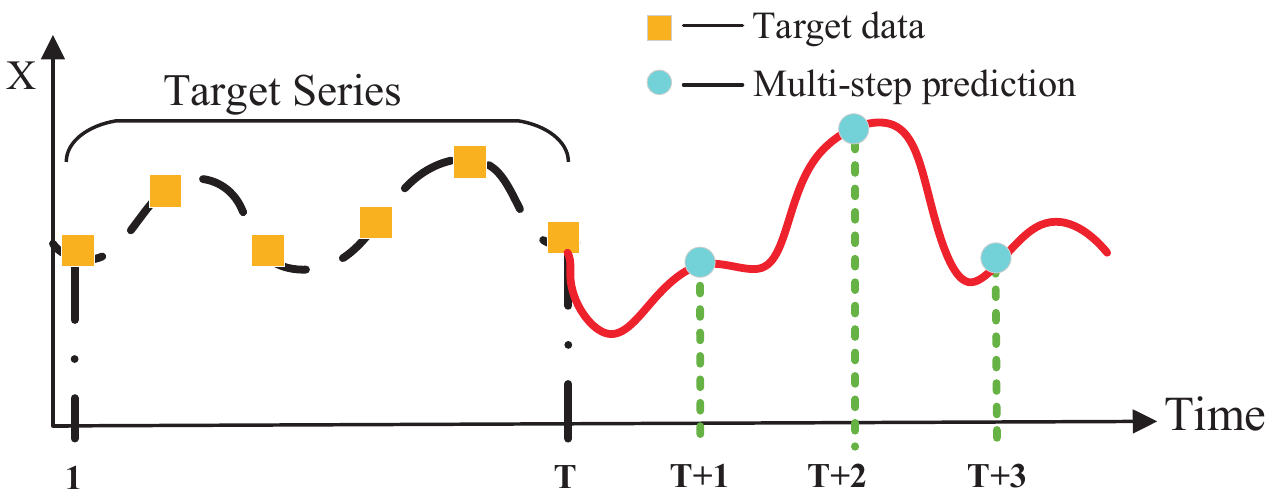}}
	\quad
	\subfigure[Arbitrary-step prediction by our continuous EgPDE-Net.]{
		\label{diff.sub1}
		\includegraphics[width=0.4\textwidth]{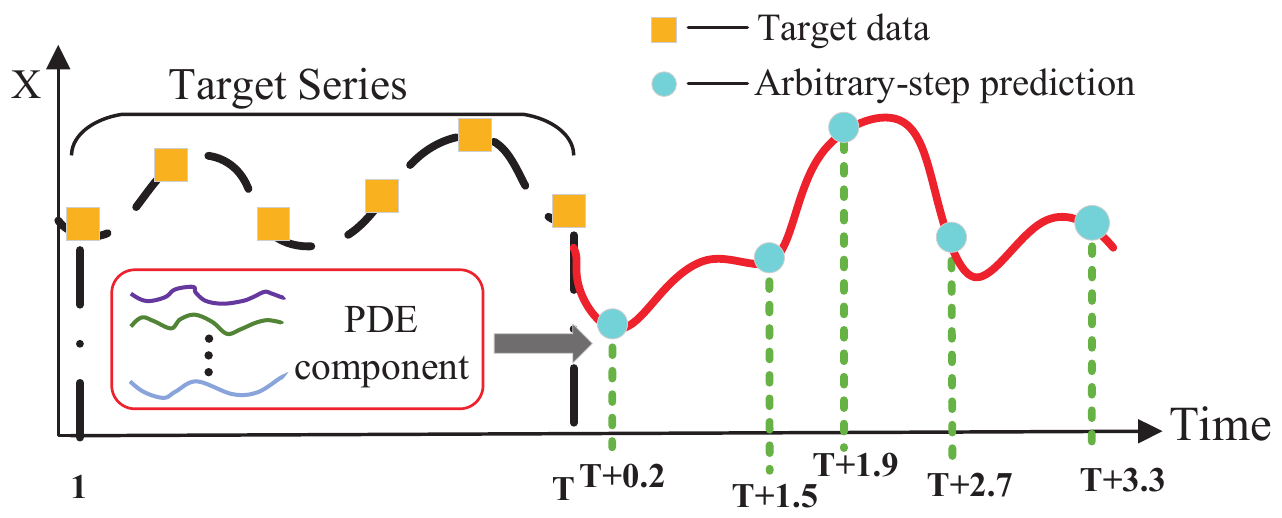}}
\caption{(a) Discrete neural networks could only make predictions at fixed time points $T+1$, $T+2$, and $T+3$, which have the same sample gap as the input data.  (b) The proposed continuous network with PDE makes predictions at continuous time points $T+M$ where $M=0.2, 1.5$, etc.
	}
	\label{diff}
\end{figure}

\IEEEPARstart{T}{ime} series analysis is an essential topic in diverse real-world scenarios, such as power prediction \cite{chen2021multiscale}, financial investment \cite{WangSLCZ19}, air quality assessment \cite{xu2019multitask}, clinical analysis \cite{mulyadi2021uncertainty},  and traffic forecasting \cite{9151256}. Accurate prediction of the future evolution helps people make important decisions for benefit maximisation.
With more demanding scenarios, one challenging and meaningful task is to forecast continuous multiple future values of one specific target series at arbitrary time points with multivariate time series.
Most deep learning structures, including Recurrent Neural Networks (RNNs), are interpreted as a discrete approximation to sequence prediction, which could only forecast the fixed step size of future values \cite{yu2017long,FoxAJPW18}. This discretisation typically breaks down for arbitrary-step prediction \cite{gao2022explainable}.
When dealing with this arbitrary-step prediction problem, a more practical approach is to build continuous models for the dynamical behaviour of multivariate data.
Fig.~\ref{diff} shows the different predictive between traditional discrete networks and our proposed continuous exogenous-variable-guided framework.

Building continuous networks has drawn much attention in academic research fields in a variety of applications such as time series analysis, node classification with graph neural network, set modelling, and normalising flows \cite{chen2018neural,rubanova2019latent,poli2019graph,xhonneux2020continuous,YBSO20,GrathwohlCBSD19}.
The neural ordinary differential equation (ODE) is one popular mainstream proposed in \cite{chen2018neural} to model time series with a continuous-time approach, in which neural networks parameterise the derivative of the hidden states.
While multivariate data are input for arbitrary-step prediction, the exogenous variables have different impacts on the target series for the prediction performance. Modelling the interaction among the exogenous variables could provide extra information and more accurate predictions. Despite the benefit, inter-series correlation and time dependence among the exogenous variables are rarely considered in the present continuous methods.

To better cope with both building continuous networks for the arbitrary-step prediction and modelling the impacts of the exogenous variables jointly, we propose a general continuous-time method called \underline{E}xogenous-\underline{g}uided \underline{P}artial \underline{D}ifferential \underline{E}quation Network (EgPDE-Net) in this paper.
The exogenous variables contain a rich data structure and information, especially the interactions among variables, which could further benefit the prediction performance. As a critical contribution, the proposed PDE framework can better model the influence of exogenous variables, which is rarely studied in the existing ODE-based methods.
Unlike current mainstreams for continuous-time modelling, i.e. the ODE based networks, EgPDE-Net is the first model built on a more general partial differential equation (PDE) framework. It describes the multivariate time series by referring to the Cauchy problem in the heat conduction equation:
\begin{equation}
    \Dot{\mathbf{z}}(y,\mathbf{x},t) := F(y,\mathbf{x},\mathbf{z},t,\bigtriangledown_x\mathbf{z},\bigtriangledown_x^2\mathbf{z},\cdots).
\end{equation}

Physically, PDEs are the fundamental equations of many important disciplines exploring the mysteries of the universe~\cite{guo2021partial}. Mathematically, they describe the changes that can be expressed continuously in the system through partial derivatives of multiple independent variables. More importantly, PDEs enjoy an elegant mechanism to model the derivatives of input multiple variables and latent states; this is more theoretically appealing than ODE systems which only describe the evolution of latent states w.r.t. time. Consequently, whilst making accurate continuous predictions, PDEs offer a better capability to model the impact of exogenous variables.

On the practical front, we design a tractable way to solve the PDE problem  for arbitrary-step prediction by utilising two ODE nets to achieve the transition between PDE and ODE. Specifically, the exogenous variables are processed with a self-attention block extracting the global and local information simultaneously. The correlation among exogenous variables could be captured in the encoding phase. The attention mechanism takes advantage of allocating different weights representing each series, which could provide interpretability to distinguish the different contributions among different driving time series \cite{heo2018uncertainty,guo2019exploring}. The first ODE network is applied to generate the partial differential weights in both the time and variable domain to guide the predicted target series trajectory. On the other hand, the target series is processed with a GRU block to capture the temporal dependence. 
The second ODE network could generate the final latent states with the hidden representation of the target series and partial differential weights.

It is noted that some multivariate time series prediction methods can implicitly model the impact of exogenous variables by forecasting all the input series simultaneously \cite{lai2018modeling,binkowski2018autoregressive,cheng2020towards}. On the one hand, these methods lack the capacity of forecasting continuous multiple future values and can only be applied to traditional discrete cases. On the other hand, they could not pinpoint the influence of exogenous variables on the specific target series.

The main contributions of this paper could be summarised as follows:
\begin{itemize}
    \item We propose a novel continuous method to consider both the intra-series temporal patterns for the target series and the inter-series correlations among the exogenous variables for multivariate time series analysis. 
    \item To the best of our knowledge, the designed general PDE framework, called EgPDE-Net, is the first work to build continuous-time representation for multivariate time series as a PDE problem. The specially designed architecture could transform the PDE problem into the ODE problem with the tool of an ODE solver, which makes the PDE problem easier to solve and feasible to conduct arbitrary-step prediction.
    \item Experiments show that EgPDE-Net performs better than the strong baselines over four multivariate time series. On average, it outperforms the best baseline by reducing $9.85\%$ on RMSE and $13.98\%$ on MAE for arbitrary-step prediction.
\end{itemize}


\section{Related Work}

\subsection{Multivariate time series}
Classical methods, including the Autoregressive (AR) model \cite{akaike1969fitting} and Vector Autoregressive model \cite{sims1980macroeconomics}, have shown their effectiveness for various real-world applications based on a linear behaviour given the past values of the series. However, the linearity limits them to model complex nonlinear characteristics in multivariate time series.
In the recent decade, deep learning methods have experienced booming development for nonlinear high-dimensional time-varying problems \cite{zhang2018varying,zhang2020mutual,zhang2021novel} with the capability of handling nonlinear problems in multi-step time series prediction task \cite{fox2018deep, yu2017long}.
In the work of \cite{zang2017deep}, the authors built a model with Long-Short Term Memory (LSTM) architecture to forecast multiple values on web traffic data. Convolution Neural Networks could also be applied for multivariate time series prediction. \cite{binkowski2018autoregressive} proposed a novel convolutional network for predicting multivariate asynchronous time series called Signi?cance-Offset Convolutional Neural Network (SOCNN). This model was designed to combine the autoregressive (AR) model and CNN. There are two convolutional parts in this model architecture. One captures the local significance of observed data, while the other represents the predictors entirely independent of position in time.
Methods in \cite{lai2018modeling, cheng2020towards} combined CNN and RNN, which aim to predict the future value of each individual variable for multivariate time series. Both of them generated forecasting values of all the series simultaneously, taking their historical data as input. These approaches have limitations in representing the contributions and influences to the target series. In contrast, our method focuses on the specific target series and makes predictions using other exogenous variables' information.

\subsection{Variant of ODE net}
With the emergence of neural ODE, researchers have begun to focus on building continuous models to solve complicated problems. The ODE net is applied for irregularly-sampled time series classification in~\cite{rubanova2019latent}. The authors added the ODE network to the loop of the RNN network. The previous hidden states were modified by an ODE network before being updated in the next step. They designed an encoder with ODE-RNN to process the irregularly-sampled data instead of fixing the sampled gap by imputation.
In the work of~\cite{KidgerMFL20}, the authors considered the influence of subsequent observations when adjusting the trajectory of the latent states. They modified the ODE net through the controlled differential equations (CDE). The derivative of the input data w.r.t. time $t$ is multiplied to the ODE function when integrating the latent states. {The authors in \cite{jhin2022exit} aimed to enhance NCDE and extended the interpolation-based NCDE model with both extrapolation and interpolation algorithms. They built another latent continuous path from a discrete time-series input by using an encoder-decoder architecture.}
In~\cite{FinlayJNO20}, the authors stated that training neural ODEs on large datasets had not been tractable due to the necessity of allowing the adaptive numerical ODE solver to refine its step size to minimal values. They proposed a theoretically-grounded combination of both optimal transport and stability regularisation to tackle this problem, which encouraged neural ODEs to prefer simpler dynamics. They added a kinetic energy regularisation term and a Jacobian Frobenius norm regularisation term to the loss function. The results show that this framework could achieve acceptable performance and great reductions in time consumed.
The existing ODE related methods rarely consider the influence of other exogenous variables, which could improve the forecasting performance. In our method, the specially designed PDE framework could utilise the impact of multivariate time series and provide a better prediction.

\subsection{Variant of PDE net}
In the objective world, PDEs are extensively applicable to describe many propagative systems modelling the relationship of the partial derivatives w.r.t. time and space, such as heat dissipation, the behaviour of sound waves, disease progression, fluid dynamics, weather patterns, or cellular kinetics \cite{courant2008methods, IakovlevHL21}. Such PDE models are considered the cornerstone of natural science and are utilised to describe most of the fundamental laws in Physics.
Recently, \cite{long2018pde} has proposed a deep feed-forward method PDE-Net to discover the hidden PDE dynamical behaviour with a convolutional neural network. Further works have extended the approach with symbolic neural networks \cite{long2019pde} and graph neural networks \cite{IakovlevHL21}. These models are designed to solve the specific dynamical systems of PDEs, which could be considered as a simulation of the underlying systems. They lack the contribution of the exogenous variables when making predictions for the target series. {The authors in \cite{luo2022learning} proposed a new learning framework to automatically model the differential operators for multivariate time series. They design a polynomial-block with convolutional layers to learn the potential derivatives. Unlike the existing PDE methods, we aim to model multivariate time series for arbitrary-step prediction, assuming that the dynamical system is governed by an unknown PDE and supports continuous evolution over time. Our method treats the driving series as a guided term and globally learns the derivatives of exogenous variables.}

\subsection{Attention mechanism in time series prediction}
Besides making predictions, providing interpretability is also important for multivariate time series models. The attention mechanism is an appropriate method to provide interpretability. Two types of attention generated by two independent RNNs are applied in~\cite{choi2016retain, heo2018uncertainty} to provide interpretable insights into the data. They leveraged two RNNs generating alpha and beta attention representing the importance of time and variables. In \cite{guo2019exploring}, the authors designed a parallel network to enhance the interpretability with a tensorized LSTM structure for single-step prediction. The network first generated temporal attention given the input data. Then variable attention is generated according to the temporal attention and the hidden states in RNN.
Self-attention is proposed in \cite{VaswaniSPUJGKP17} for language modelling. The authors developed a transformer framework by using an Encoder-Decoder structure. The network generates query, key and value given the input data. Then a softmax operation is conducted on the matrix multiplication of query and key to generate the attention weights. The self-attention in the transformer could help the network focus on more relevant latent states at specific time steps.
Another work of \cite{zhou2020informer} improved the transformer architecture by proposing a ProbSparse Self-attention mechanism which allows each key only to interact with the  dominant queries. This framework could leverage the most important queries, reducing the network parameters.

\section{Methodology}
In this section, we will introduce the problem statement of arbitrary-step prediction and model details of our EgPDE-Net. We leverage two ODE nets dealing with exogenous variables and target series respectively with two pipelines.
\begin{figure*}[!htbp]
    \centering
    \includegraphics[width=0.7\textwidth]{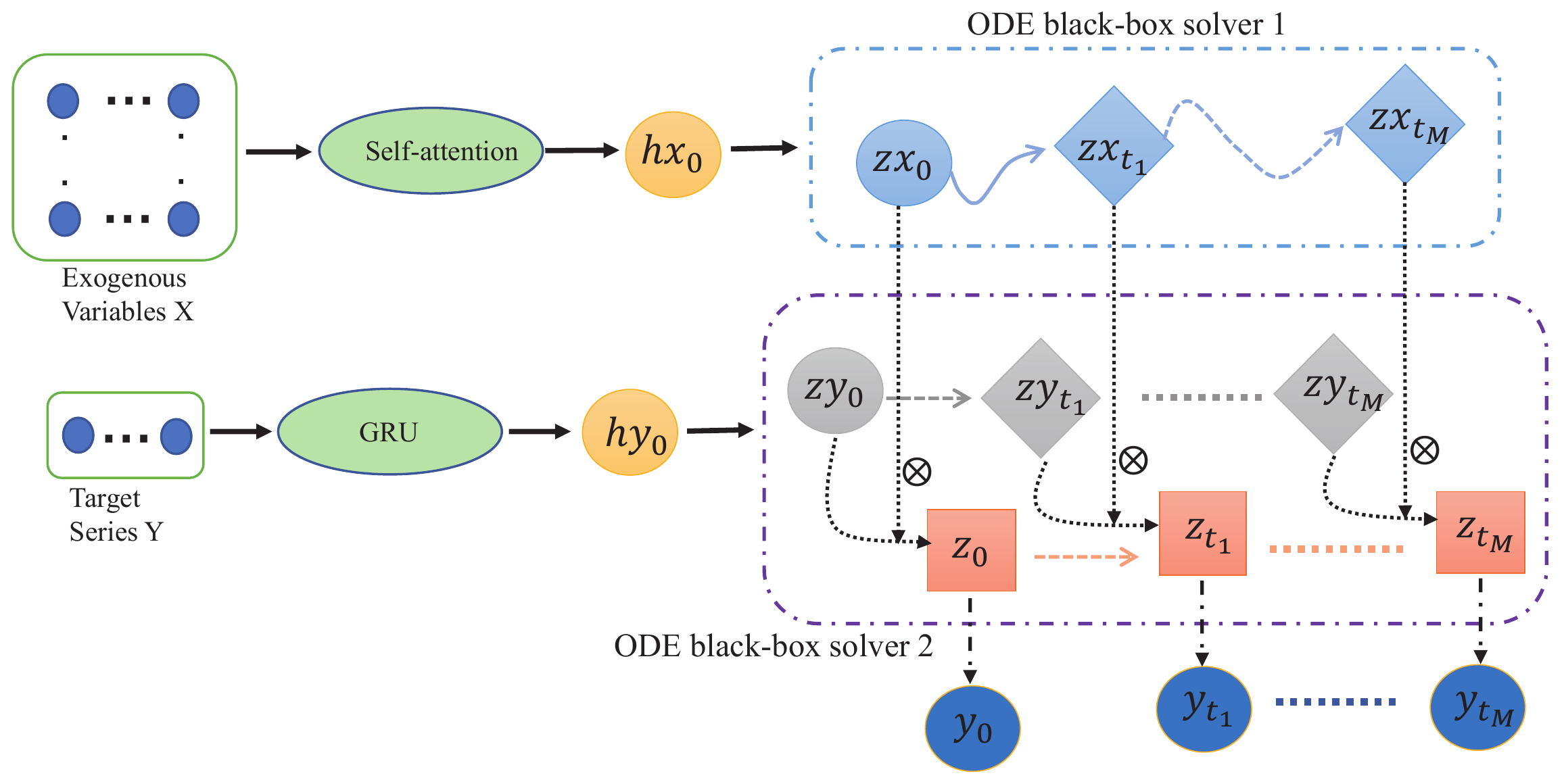}
    \caption{Model structure of EgPDE-Net. The hidden representations of $hx_0$ and $hy_0$ are generated with the self-attention block and GRU block respectively. Then the above ODE net is leveraged to obtain the latent states representing the partial derivative weights guiding the generation of the trajectory of final latent state $\mathbf{z}_t$ in the below ODE net. $t_1,\cdots,t_M$ are the forecasting time points.}
    \label{struc}
\end{figure*}
As shown in Fig.~\ref{struc}, the exogenous variables are processed with a self-attention block generating a summarised hidden representation $hx_0$. The first ODE net is used to obtain the regularised partial derivatives weights of the hidden representation w.r.t. $X_T$. The hidden representation $hy_0$ is generated using a GRU block with input of target series. The second ODE net is leveraged combining the hidden representation of target series and the regularised partial derivatives weights to guide the generation of the final latent states $\mathbf{z}_t$.

\subsection{Arbitrary-step prediction}
In time series analysis, most deep learning models aim to forecast the future value of time $T+1$ given the historical data of previous $T$ time steps. However the effective decision often requires forecasting multiple future values for multivariate time series data in many real-world problems. For instance, knowing the demand for electricity in the next few hours could help to devise a better energy use plan, and forecasting the stock market in the near or distant future could produce more profits \cite{cheng2020towards}.
This work aims to forecast the multiple future values at arbitrary time points by building a continuous model. We follow the work of \cite{gao2022explainable} for the definition of arbitrary-step prediction.

Given a multivariate time series, it consists of target series that we want to predict and exogenous variables that impact the target series in the predicting decision. With the length $T$ of historical data, the aim is to forecast the multiple future values of the target series represented as
\begin{equation}
\left[\hat{y}_{T+m_1},\cdots,\hat{y}_{T+m_K}\right] = \mathcal{F}(\mathbf{X}_T, \mathbf{Y}_T),
\end{equation}
where $K$ is the number of predicted future values. The forecasting time interval $\{T+m_1,\cdots,T+m_K\}$ could be set as any continuous value, e.g. $\{T+0.8,\cdots,T+2.2\}$. Here $\mathcal{F}(\cdot)$ is achieved by the proposed continuous framework EgPDE-Net. $\mathbf{X}_T$ is the exogenous variables input data of length $T$, and $\mathbf{Y}_T$ is the target series input data denoted as
\begin{eqnarray}
    \mathbf{X}_T &=& \{ \mathbf{x}_1,\mathbf{x}_2,\cdots,\mathbf{x}_T \} \in \mathbb{R}^{T\times N}, \\
    \mathbf{Y}_T &=& [y_1, y_2,\cdots,y_T]^\top \in \mathbb{R}^T,
\end{eqnarray}
where $\mathbf{x}_t = [x^1_t,x^2_t,\cdots,x^N_t]^\top, t = 1,\cdots,T$ and $N$ is the number of exogenous variables.


\subsection{Neural ODE net}
Neural ODE proposed in \cite{chen2018neural} is a continuous-time model to overcome the limitations of requiring discrete observation and emission intervals in RNNs. A hidden state $\mathbf{h}$ in an RNN is modelled as:
\begin{equation}
    \mathbf{h}_{t+1}=\mathbf{h}_t + f(\mathbf{h}_t,\theta_t). \label{rnn-hid}
\end{equation}
When adding more layers and taking smaller steps in Eq.~\ref{rnn-hid}, the continuous dynamics of hidden units are parameterised through an ODE specified by a neural network:
\begin{equation}
    \frac{d\mathbf{h}(t)}{dt} = f_\theta(\mathbf{h}(t), t).  \label{ode-ori}
\end{equation}
In this form, the ODE network parameterises the derivative of the hidden states w.r.t. time $t$ with parameters $\theta$ rather than directly parameterising the hidden states. 
The hidden states $\mathbf{h}(t)$ could be evaluated at desired time points by integrating the ODE function over the specific time interval with an initial value such as
\begin{equation}
    h(t_{end}) = h(t_{start}) + \int ^{t_{end}}_{t_{start}} f_\theta(\mathbf{h}(s),s)ds.
\end{equation}
Neural ODE models the incremental changes in time series, bringing more smooth and accurate estimation for prediction tasks, requiring a constant memory cost without storing any intermediate quantities of the forward pass.

\subsection{Reducing PDE problem with Regularised Guided ODE}\label{PRODE}
For some physics, chemistry, and biology problems, it is necessary to establish various mathematical models. Most of them are described by Reaction-Diffusion Equation for quantitative or qualitative analysis. Reaction-Diffusion Equation could be derived from many natural phenomena such as heat conduction in physics, substance concentration change in the chemical reaction, and species invasion process in biology.
{The matrix-valued functions $A(U(x,t))$, $B(U(x,t))$ can be used to define the elliptic operator $\mathcal{O}_L$ according to~\cite{guo2021partial,Wells2008}:
\begin{equation}
    \mathcal{O}_LU = -\nabla(A(U(x))\nabla U) + B(U(x))\nabla U.
    \label{eq8}
\end{equation}
Matrix-valued functions $A$ and $B$ can be considered the coefficient functions and are determined according to the characteristics of the specific problems. They are weight matrices to represent the PDE system. We will use the designed networks to estimate these weight matrices.
}
In mathematics, Laplace operator is expressed as:
\begin{equation}
    \nabla^2U =: \Delta U,  \quad \Delta U= \sum\nolimits_i \frac{\partial^2}{\partial x_i^2}U.
    \label{eq9}
\end{equation}

In some cases, it is meaningful to apply the relatively physical equation to the existing neural methods to improve neural network model performance. The Cauchy problem of the one-dimensional heat conduction equation $f(x,t)$ with the constrained mapping $\varphi(\cdot)$ is expressed as follows:
\begin{equation}
\begin{array}{cc}
    \mathbf{u}_t - a^2\mathbf{u}_{xx} = f(x,t),  -\infty < x <  +\infty, t > 0,&\\
    \mbox{subject \ to} \  \mathbf{u}|_{t=0} = \varphi(x),-\infty < x <  +\infty,
    \end{array}
    \label{eq1}
\end{equation}
where $u_t$ is the first order derivative w.r.t. time $t$, and $u_{xx}$ is the second order derivative w.r.t. variable $x$.
Many problems could be modelled concerning the heat conduction equation in real-world applications. In time series analysis, the sequence changes could be considered as a delivery process similar to the diffusion equation. The future values of the sequence will converge with time $t$ evolving given the historical data.
When processing  with multivariate time series, the target series is influenced by the historical information of itself and other exogenous variables. Inspired from the idea of general neural PDE in the work of~\cite{guo2021partial}, the PDE problem for multivariate time series analysis could be defined as follows:
{
\begin{equation}
\small
    \partial \mathbf{z}/\partial t=\mathcal{F}(y,x_1,\cdots,x_n,t,\frac{\partial^2 \mathbf{z}}{\partial x_1^2},\cdots,\frac{\partial^2 \mathbf{z}}{\partial x_n^2},\mathbf{z},A(\mathbf{z}),B(\mathbf{z})),
    \label{eqpde}
\end{equation}}
where $\mathbf{z}$ represents the trajectory of target series in latent space, $x_i$ represents the individual exogenous series, and $A(\mathbf{z}),B(\mathbf{z})$ are the function of latent state $\mathbf{z}$.

The defined nonlinear PDE problem is complicated and usually has no explicit closed-form solution. Inspired from the previous works~\cite{bao2020numerical,beck2021deep,sirignano2018dgm,chen2020friedrichs,han2017deep,IakovlevHL21} approximating the PDEs with weak solutions, we reduce the PDE problem to an exogenous-guided ODE problem parameterised by using two neural networks as the weak solutions.
The complicated PDE problem could be transformed into a simpler modelling problem and is easier to get the numerical results which also improves the forecasting performance with the weak solutions.
In this case, we define the following lemma to transform the PDE problem into a regularised exogenous-guided ODE problem:
\begin{lem}
For simplicity, we use $\nabla^2$ to represent  $\sum_i \frac{\partial^2 \mathbf{z}}{\partial \mathbf{x}_i^2}$ according to Eq.~\ref{eq9}.
In Eq.~\ref{eqpde}, referring to the reaction-diffusion equations and quasi-linear approximations for multivariate time series by leveraging Eq.~\ref{eq8}, {the defined PDE problem of Eq.~\ref{eqpde} could be substituted with:
\begin{equation}
\begin{split}
    &\frac{\partial \mathbf{z}}{\partial t} = \mathcal{O}_L\mathbf{z} = -\nabla(A(\mathbf{z})\nabla{\mathbf{z}}) + B(\mathbf{z})\nabla\mathbf{z}.\label{eqOL}
\end{split}
\end{equation}
Here, $A(\mathbf{z})$ is the diffusion matrix and $B(\mathbf{z})\nabla\mathbf{z}$ is the convection vector. Then we rewrite the Eq.~\ref{eqOL} according to the differential criterion as follows:
\begin{equation}
\begin{split}
    \frac{\partial \mathbf{z}}{\partial t} &= -A(\mathbf{z})\nabla^2\mathbf{z}(t) - \nabla{A(\mathbf{z})}\nabla\mathbf{z}(t) + B(\mathbf{z})\nabla\mathbf{z}(t)\\
    &= -A(\mathbf{z})\nabla^2\mathbf{z}(t) + (B(\mathbf{z}) - \nabla{A(\mathbf{z})})\nabla\mathbf{z}(t).
\end{split}
\end{equation}
We merge $\nabla{A(\mathbf{z})}$ and $B(\mathbf{z})$ into new parameter $B'(\mathbf{z})=B(\mathbf{z}) - \nabla{A(\mathbf{z})}$. Then we could have the representation of $\partial\mathbf{z}/\partial t$ as follows:
\begin{equation}
    \frac{\partial \mathbf{z}}{\partial t} = -A(\mathbf{z})\nabla^2\mathbf{z}(t) + B'(\mathbf{z})\nabla\mathbf{z}(t).
    \label{pdezt}
\end{equation}
When we take the exponential operation on both side of Eq.~\ref{pdezt}, we could have the variant of the derivative of $\mathbf{z}$:
\begin{equation}
\begin{split}
    exp(\frac{\partial \mathbf{z}}{\partial t}) &= exp(-A(\mathbf{z})\nabla^2\mathbf{z}(t) + B'(\mathbf{z})\nabla\mathbf{z}(t))\\
    &= exp(-A(\mathbf{z})\nabla^2\mathbf{z}(t))\cdot exp(B'(\mathbf{z})\nabla\mathbf{z}(t)).\label{expzt}
\end{split}
\end{equation}
The two parts of the right side of Eq.~\ref{expzt} are estimated with two neural networks $G(\cdot)$ and $f(\cdot)$:
\begin{align}
           &exp(-A(\mathbf{z})\nabla^2\mathbf{z}(t)) =: G(\mathbf{z}_\mathbf{X}), \\
           &exp(B'(\mathbf{z})\nabla\mathbf{z}(t)) =: f(\mathbf{z}_\mathbf{Y}).
\end{align}
With the approximation and transition by the two neural networks, we could obtain the partial derivative of the latent states $\mathbf{z}$:
\begin{equation}
    \frac{\partial \mathbf{z}}{\partial t} =: ln(G(\mathbf{z}_\mathbf{X})*f(\mathbf{z}_\mathbf{Y})),
    \label{odefunc}
\end{equation}
which is an example of Neural ODE \cite{chen2018neural}.
Briefly, we name the ODE in Eq.~\ref{odefunc} the exogenous guidance ODE, which could be numerically solved using the black-box ODE solver.}
\end{lem}

\begin{figure}
    \centering
    \includegraphics[width=0.45\textwidth]{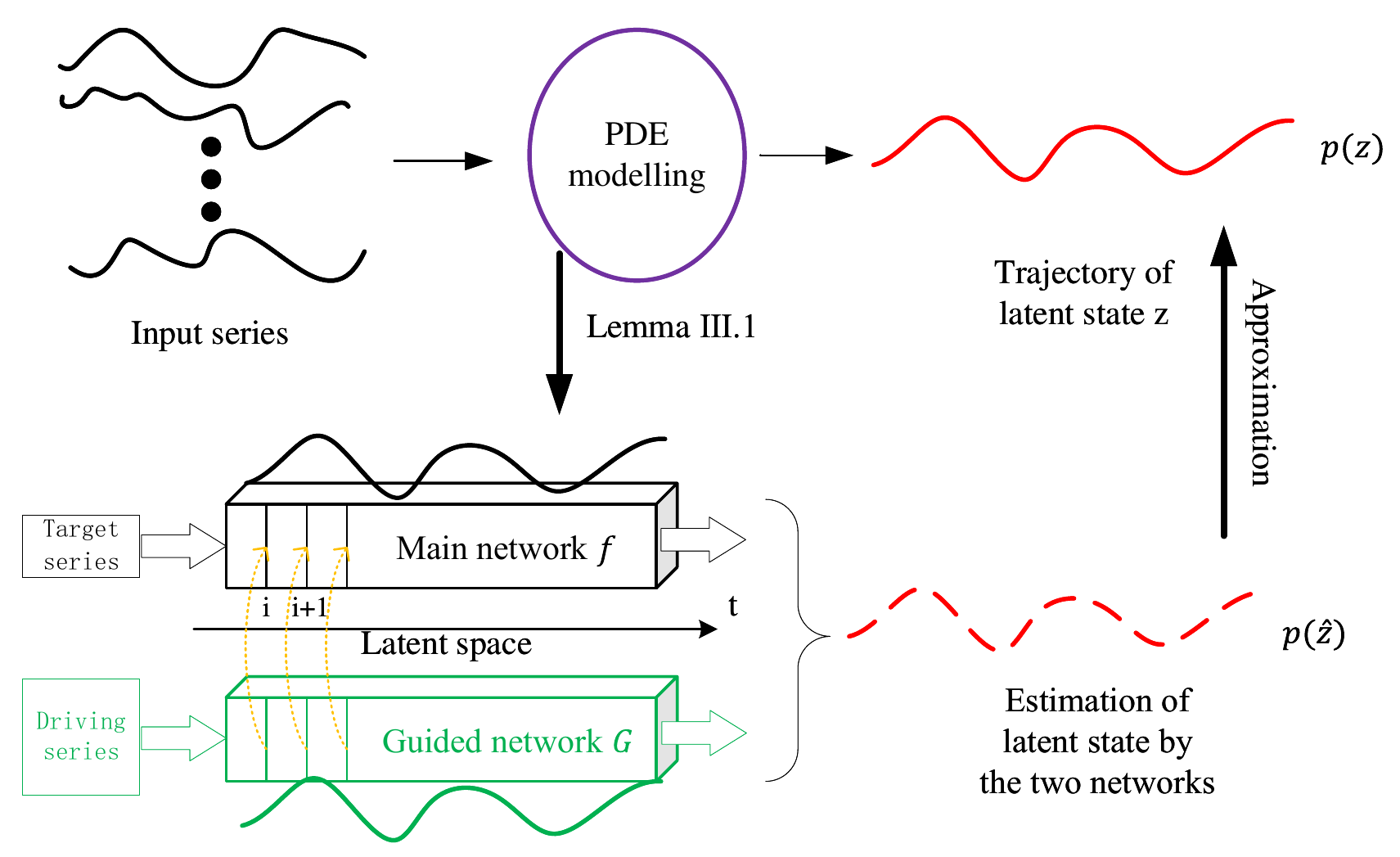}
    \caption{Graphic illustration of the transition of Lemma \uppercase\expandafter{\romannumeral3}.1}
    \label{lemma}
\end{figure}
The function $G(\cdot)$ is achieved with the first ODE network to get the regularised partial representation of exogenous variables.
In this task, the target series is influenced by many exogenous variables. The element-wise product is used in Eq.~\ref{odefunc} and $G(\cdot)$ is like a weight matrix representing the weight ratio that affects and guides the generated trajectory of the target series for final prediction in the latent space. The first ODE net is applied as follows:
\begin{align}
     &\mathbf{z}_x = G(\mathbf{z}(\mathbf{X})) = \text{ODESolve}(\mathbf{z}_{x_0},\theta_g,t), \\
     &\mathbf{z}_{x_0} = \text{Attblock}(\mathbf{X}_T).
\label{ode1}
\end{align}
In Eq.~\ref{ode1}, $\text{Attblock}(\cdot)$ is achieved by a self-attention block, which could capture the temporal connection and variable-wise correlation among the exogenous variables. The attention weights could be learnt automatically in the self-attention block.
In Eq.~\ref{odefunc}, $f$ is parameterised with another neural network and $G$ is the regularised term to adjust the ODE function in generation phrase. The final latent states $\mathbf{z}_t$ are generated with the second ODE net conditioned on $\mathbf{z}_x$.
\begin{align}
     &\mathbf{z}_{y_0} = \text{GRU}(\mathbf{Y}), \\
     &\mathbf{z}_t = \text{ODESolve}(\mathbf{z}_{y_0},\mathbf{z}_x,\theta_f,t).
\end{align}
For $t \in (t_0,t_n]$, the solution of $\mathbf{z}_{t_n}$ given the initial value $\mathbf{z}_{t_0}$ could be computed as
\begin{equation}
    \mathbf{z}_{t_n} = \mathbf{z}_{t_0} + \int ^{t_n}_{t_0}ln(G_{\theta_g}(\mathbf{z}_{\mathbf{X}}) [f_{\theta_f}({\mathbf{z}_{y}}_s)])ds.
    \label{eq10}
\end{equation}
It is tractable to solve the PDE problem using the same techniques as for Neural ODE. In the experiments, we use the existing torchdiffeq package~\cite{chen2018neural} with modifying the computation of the ODE function.

\subsection{Self-attention block}
The basic self-attention proposed in~\cite{VaswaniSPUJGKP17} leverages the scaled dot-product to compute the attention with the tuple input (query, key, value) derived from raw data:
\begin{equation}
    \text{Attention}(Q, K, V) = \text{softmax}(\frac{QK^\top}{\sqrt{d_k}})V,
\end{equation}
where $d_k$ is the dimension of one query or key. The softmax operation is used to normalise the weights into a probability distribution.
Matrices $Q$, $K$ and $V$ are the hidden representations of raw exogenous variables parameterised with neural networks.

Instead of using one single $d_{model}$-dimensional attention function, it is beneficial to perform multi-head attention by linearly projecting the queries, keys and values $h$ times with differently learned linear projections. Multi-head attention could allow the model to capture information from different representation subspaces at different positions jointly. The self-attention block has an attractive capability of capturing the intra-relationship in a single sequence and the inter-correlation among different sequences.
\begin{equation}
\small
    \begin{array}{lc}
         \text{MultiHead}_\text{Att}(Q,K,V) = \text{Concat}(head_1,\cdots,head_h)W^O,&  \\
         \mbox{where} \  head_i = \text{Attention}(QW_i^Q,KW_i^K,VW_i^V),
    \end{array}
\end{equation}
with parameter matrices $W_i^Q \in \mathbb{R}^{d_{model}\times d_k}, W_i^K \in \mathbb{R}^{d_{model}\times d_k},W_i^V \in \mathbb{R}^{d_{model}\times d_v}$ and $W^O \in \mathbb{R}^{hd_v\times d_{model}}$.


\subsection{Loss function}\label{loss_func}
The objective function is the Mean Square Error (MSE) between the predicted values and the true values:
\begin{equation}
\small
    \mathcal{L}_{mse} = \frac{1}{L}\sum\nolimits^L_{i=1} \frac{1}{K} \sum\nolimits^K_{t=1} (\hat{y}^i_t - y^i_t)^2, \label{loss}
\end{equation}
where $L$ is the number of training samples and $K$ is the number of predicted future values. The goal in the training stage is to minimise the loss function given the parameter set of our model by using gradient descent methods for optimisation, such as the Adam algorithm~\cite{KingmaB14}.

\subsection{Complexity Analysis}
Assume that the hidden size of the self-attention block is $d_{model}$ and the number of exogenous variables is $N$, the attention module processing the exogenous variables has computation complexity $d_{model}^2 + N\cdot d_{model}$. As for the update process of the ode function, the complexity is $d_{rnn}^2$, where $d_{rnn}$ is the hidden dimension of the ode function. Overall, the computation complexity is $\mathcal{O}(d_{model}^2 + N\cdot d_{model} + d_{rnn}^2)$. In general, though it may take more parameters and computations, the extra overhead does not affect our model¡¯s application in real scenarios. More importantly, our method could model the impacts of the exogenous variables effectively and lead to significant improvements in performance. On average, our method outperforms the best baseline by reducing $9.85\%$ on RMSE and $13.98\%$ on MAE for arbitrary-step prediction. {Taking the Electricity dataset as one illustrative example, we demonstrate the number of parameters, the training time for one epoch, and the testing time on the Electricity dataset in Table~\ref{tab:parameter}. The results show that our method has a competitive inference time with a relatively small number of parameters.}

\begin{table}[h]
\centering
\caption{{Empirical model efficiency comparisons among the competitive methods. We list the number of parameters, the training time for one epoch, and the testing time on the Electricity dataset.}}
\label{tab:parameter}
\begin{tabular}{c|ccc}
\toprule
Model          & \# of parameters & train time (s) & test time (s) \\ \hline
LSTNet         & 72,175            & 1.82           & 0.20          \\
Latent ODE-RNN & 62,541            & 43.19          & 2.08          \\
MTGODE         & 70,901            & 55.45          & 2.46          \\
STG-NCDE       & 2,543,301          & 171.49         & 6.19          \\
STGODE         & 610,741           & 24.75          & 0.77          \\
ETN-ODE        & 7,302             & 33.36          & 1.64          \\
EgPDE-Net      & 50,859            & 55.76          & 1.85          \\ \bottomrule
\end{tabular}
\end{table}

\begin{table}[h]
\centering
\begin{tabular}{|c|c|c|c|c|}
\hline
Dataset     & SML2010   & Electricity & ETTh1  & ETTh2 \\ \hline\hline
\#Instances & 4,137    & 22,201      & 17,420  & 17,420   \\ \hline
\#Features  & 13          & 15          & 6     & 6     \\ \hline
Sample rate & 1min        & 1h          & 1h    & 1h    \\ \hline
Train size  & 80\%       & 80\%        & 80\%   & 80\%  \\ \hline
Valid size  & 10\%       & 10\%        & 10\%   & 10\%  \\ \hline
Test size   & 10\%       & 10\%        & 10\%   & 10\%  \\ \hline
\end{tabular}
\caption{Summary of four datasets}
\label{data}
\end{table}
\section{Experiments}
In this section, we conduct extensive experiments on four real-world datasets for arbitrary-step and standard multi-step predictions of multivariate time series against three continuous-time methods and two non-continuous methods. {The code is available at \url{https://github.com/PengleiGao/EgPDE-net}.}

\subsection{Datasets}
As summarised in Table~\ref{data}, the four datasets are publicly available. {The train/validation/test sets are obtained with a split ratio of 8:1:1 following the previous works \cite{gao2022explainable,guo2019exploring}.}
\begin{itemize}
\item \textbf{SML2010}~\cite{zamora2014line}:
It is a public dataset used for indoor temperature forecasting sampled every minute. The room temperature is taken as the target series, and another $13$ time series are exogenous variables containing approximately $40$ days of monitoring data.
\item \textbf{Electricity}~\cite{elec-data}:
This is a public dataset for electricity consumption prediction of Homestead, US. The consumption is chosen as the target series sampled hourly, while the other $15$ time series are exogenous variables containing weather features.
\item \textbf{Electricity Transformer Temperature (ETT)}~\cite{zhou2020informer}:
ETT is a crucial indicator in electric power deployment.
This dataset collected two-year data from two separated counties in China. The data are sampled hourly. According to the two counties, it consists of two separated datasets as $\{$ ETTh1, ETTh2 $\}$. There are 6 power load features, and the oil temperature is selected as the target series.
\end{itemize}

\subsection{Experimental settings}
For all the datasets, the window size $T$ is chosen as 20 following the baseline methods. Our proposed model EgPDE-Net is implemented in PyTorch with a mini-batch size of 128, and a learning rate of 0.001 for SML2010 and 0.01 for the other datasets. For EgPDE-Net, the hidden size of RNN is chosen from \{32, 64, 128\}, and the dimension of self-attention is chosen from \{15, 30, 45, 60\}. The number of heads in self-attention is chosen from \{2, 4, 6, 8\}.
Each method is trained five times to report the average performance for comparison. Two standard evaluation metrics are chosen for the prediction task to evaluate the performance of all the methods.
\begin{itemize}
\small
    \item Root Mean Squared Error:
    $\mbox{RMSE} = \sqrt{\frac{1}{p} \sum^p_{t=1} (\hat{y}_t - y_t)^2}$.
    \item Mean Absolute Error:
    $\mbox{MAE} = \frac{1}{p} \sum^p_{t=1} |\hat{y}_t - y_t|$.
\end{itemize}

\begin{table*}[t]
\centering
\caption{RMSE of each step for arbitrary-step prediction on different datasets.}
\scalebox{0.83}{
\begin{tabular}{ccccccc}
\toprule\hline
                                    & Step1                & Step1.5              & Step2                & Step2.5              & Step3                & Average              \\ \cline{2-7}
\multicolumn{7}{c}{SML2010}                                                                                                                                                   \\ \hline
\multicolumn{1}{c|}{LSTNet}     & 0.452$\pm$0.086          & -          & 0.625$\pm$0.557          & -          & 0.426$\pm$0.051          & 0.546$\pm$0.242          \\
\multicolumn{1}{c|}{IMV-tensor}     & 0.278$\pm$0.086          & -          & 0.313$\pm$0.113          & -          & 0.382$\pm$0.096          & 0.328$\pm$0.093          \\
\multicolumn{1}{c|}{Latent-ODE}     & 0.851$\pm$0.173          & 0.877$\pm$0.172          & 0.905$\pm$0.168          & 0.935$\pm$0.167          & 0.967$\pm$0.164          & 0.908$\pm$0.168          \\
\multicolumn{1}{c|}{Latent ODE-RNN} & 0.385$\pm$0.059          & 0.411$\pm$0.047          & 0.441$\pm$0.035          & 0.478$\pm$0.023          & 0.521$\pm$0.017          & 0.450$\pm$0.032           \\
\multicolumn{1}{c|}{ETN-ODE}        & 0.146$\pm$0.028          & 0.156$\pm$0.020           & 0.175$\pm$0.015          & 0.199$\pm$0.011          & 0.228$\pm$0.009          & 0.183$\pm$0.011          \\
\multicolumn{1}{c|}{\textbf{EgPDE-Net}} & \textbf{0.099$\pm$0.007} & \textbf{0.120$\pm$0.007} & \textbf{0.145$\pm$0.008} & \textbf{0.171$\pm$0.010} & \textbf{0.200$\pm$0.012}  & \textbf{0.151$\pm$0.008} \\ \hline\bottomrule
\multicolumn{7}{c}{ETTh1}                                                                                                                                                     \\ \hline
\multicolumn{1}{c|}{LSTNet}     & 1.138$\pm$0.043          & -          & 1.214$\pm$0.022          & -          & 1.383$\pm$0.033          & 1.249$\pm$0.030          \\
\multicolumn{1}{c|}{IMV-tensor}     & 1.017$\pm$0.069          & -          & 1.264$\pm$0.049          & -          & 1.444$\pm$0.040          & 1.254$\pm$0.044          \\
\multicolumn{1}{c|}{Latent-ODE}     & 1.291$\pm$0.113          & 1.374$\pm$0.104          & 1.483$\pm$0.103          & 1.537$\pm$0.104          & 1.631$\pm$0.110           & 1.468$\pm$0.106          \\
\multicolumn{1}{c|}{Latent ODE-RNN} & 1.836$\pm$0.403          & 1.869$\pm$0.403          & 1.955$\pm$0.392          & 1.980$\pm$0.394           & 2.056$\pm$0.387          & 1.941$\pm$0.395          \\
\multicolumn{1}{c|}{ETN-ODE}        & 1.026$\pm$0.036          & 1.108$\pm$0.027          & 1.200$\pm$0.036            & 1.271$\pm$0.041          & 1.372$\pm$0.040           & 1.202$\pm$0.031          \\
\multicolumn{1}{c|}{\textbf{EgPDE-Net}} & \textbf{0.863$\pm$0.009} & \textbf{1.028$\pm$0.012} & \textbf{1.162$\pm$0.011} & \textbf{1.263$\pm$0.015} & \textbf{1.372$\pm$0.020}  & \textbf{1.151$\pm$0.013} \\ \hline\bottomrule
\multicolumn{7}{c}{ETTh2}                                                                                                                                                     \\ \hline
\multicolumn{1}{c|}{LSTNet}     & 2.043$\pm$0.162          & -          & 2.487$\pm$0.073          & -          & 3.896$\pm$0.124          & 2.920$\pm$0.046          \\
\multicolumn{1}{c|}{IMV-tensor}     & 1.591$\pm$0.105          & -          & 2.587$\pm$0.158          & -          & 3.327$\pm$0.127          & 2.601$\pm$0.122          \\
\multicolumn{1}{c|}{Latent-ODE}     & 4.173$\pm$0.742          & 4.489$\pm$0.673          & 4.749$\pm$0.605          & 5.042$\pm$0.531          & 5.246$\pm$0.483          & 4.759$\pm$0.581          \\
\multicolumn{1}{c|}{Latent ODE-RNN} & 3.745$\pm$0.407          & 3.965$\pm$0.347          & 4.109$\pm$0.299          & 4.300$\pm$0.277            & 4.400$\pm$0.266            & 4.112$\pm$0.304          \\
\multicolumn{1}{c|}{ETN-ODE}        & 1.620$\pm$0.181           & 2.103$\pm$0.143          & 2.523$\pm$0.142          & 2.974$\pm$0.149          & 3.338$\pm$0.159          & 2.585$\pm$0.139          \\
\multicolumn{1}{c|}{\textbf{EgPDE-Net}} & \textbf{1.272$\pm$0.017} & \textbf{1.771$\pm$0.015} & \textbf{2.222$\pm$0.022} & \textbf{2.680$\pm$0.029}  & \textbf{3.005$\pm$0.039} & \textbf{2.276$\pm$0.023} \\ \hline\bottomrule
\multicolumn{7}{c}{Electricity}                                                                                                                                               \\ \hline
\multicolumn{1}{c|}{LSTNet}     & 7.530$\pm$0.472          & -          & 5.505$\pm$0.216          & -          & 10.789$\pm$0.52          & 8.242$\pm$0.161          \\
\multicolumn{1}{c|}{IMV-tensor}     & 3.618$\pm$0.307          & -          & 4.939$\pm$0.202          & -          & 5.728$\pm$0.194          & 4.843$\pm$0.183          \\
\multicolumn{1}{c|}{Latent-ODE}     & 12.964$\pm$1.494         & 13.245$\pm$1.668         & 13.634$\pm$1.950          & 14.069$\pm$2.300           & 14.534$\pm$2.656         & 13.707$\pm$2.004         \\
\multicolumn{1}{c|}{Latent ODE-RNN} & 7.720$\pm$1.503           & 7.961$\pm$1.712          & 8.176$\pm$1.983          & 8.454$\pm$2.254          & 8.646$\pm$2.482          & 8.204$\pm$2.004          \\
\multicolumn{1}{c|}{ETN-ODE}        & 3.449$\pm$0.173          & 4.142$\pm$0.129          & 4.596$\pm$0.137          & 5.043$\pm$0.124          & 5.302$\pm$0.059          & 4.555$\pm$0.111          \\
\multicolumn{1}{c|}{\textbf{EgPDE-Net}} & \textbf{3.036$\pm$0.074} & \textbf{3.862$\pm$0.167} & \textbf{4.358$\pm$0.176} & \textbf{4.819$\pm$0.178} & \textbf{5.084$\pm$0.143} & \textbf{4.294$\pm$0.145} \\ \hline\bottomrule
\end{tabular}}
\label{tab-arbi-1}
\end{table*}

\begin{table*}[ht]
\centering
\caption{MAE of each step for arbitrary-step prediction on different datasets.}
\scalebox{0.83}{
\begin{tabular}{ccccccc}
\toprule\hline
                                    & Step1                & Step1.5              & Step2                & Step2.5              & Step3                & Average              \\ \cline{2-7}
\multicolumn{7}{c}{SML2010}                                                                                                                                                                                       \\ \hline
\multicolumn{1}{c|}{LSTNet}     & 0.380$\pm$0.070          & -          & 0.506$\pm$0.483          & -          & 0.329$\pm$0.039          & 0.405$\pm$0.132          \\
\multicolumn{1}{c|}{IMV-tensor}     & 0.218$\pm$0.070          & -          & 0.247$\pm$0.087          & -          & 0.290$\pm$0.071          & 0.252$\pm$0.073          \\
\multicolumn{1}{c|}{Latent-ODE}     & 0.679$\pm$0.149               & 0.702$\pm$0.146                 & 0.725$\pm$0.141               & 0.751$\pm$0.137                 & 0.775$\pm$0.135               & 0.727$\pm$0.141                 \\
\multicolumn{1}{c|}{Latent ODE-RNN} & 0.310$\pm$0.040                 & 0.332$\pm$0.031                 & 0.356$\pm$0.019               & 0.385$\pm$0.012                 & 0.419$\pm$0.012               & 0.360$\pm$0.019                  \\
\multicolumn{1}{c|}{ETN-ODE}        & 0.115$\pm$0.026               & 0.121$\pm$0.017                 & 0.133$\pm$0.012               & 0.149$\pm$0.005                 & 0.170$\pm$0.011                & 0.138$\pm$0.010                  \\
\multicolumn{1}{c|}{\textbf{EgPDE-Net}} & \textbf{0.074$\pm$0.006}      & \textbf{0.086$\pm$0.004}        & \textbf{0.102$\pm$0.005}      & \textbf{0.119$\pm$0.006}          & \textbf{0.142$\pm$0.006}       & \textbf{0.105$\pm$0.005}        \\ \hline\bottomrule
\multicolumn{7}{c}{ETTh1}                                                                                                                                                                                         \\ \hline
\multicolumn{1}{c|}{LSTNet}     & 0.851$\pm$0.040          & -          & 0.878$\pm$0.021          & -          & 0.997$\pm$0.025          & 0.909$\pm$0.026          \\
\multicolumn{1}{c|}{IMV-tensor}     & 0.755$\pm$0.086          & -          & 0.937$\pm$0.064          & -          & 1.067$\pm$0.079          & 0.920$\pm$0.070          \\
\multicolumn{1}{c|}{Latent-ODE}     & 0.961$\pm$0.156               & 1.022$\pm$0.123                 & 1.107$\pm$0.129               & 1.151$\pm$0.123                 & 1.219$\pm$0.129               & 1.096$\pm$0.128                 \\
\multicolumn{1}{c|}{Latent ODE-RNN} & 1.452$\pm$0.390                & 1.474$\pm$0.391                 & 1.540$\pm$0.380                 & 1.561$\pm$0.379                 & 1.623$\pm$0.372               & 1.530$\pm$0.382                  \\
\multicolumn{1}{c|}{ETN-ODE}        & 0.766$\pm$0.045               & 0.802$\pm$0.031                 & 0.878$\pm$0.037               & 0.918$\pm$0.036                 & 1.007$\pm$0.036               & 0.874$\pm$0.032                 \\
\multicolumn{1}{c|}{\textbf{EgPDE-Net}} & \textbf{0.594$\pm$0.014}      & \textbf{0.705$\pm$0.015}        & \textbf{0.823$\pm$0.017}      & \textbf{0.890$\pm$0.022}         & \textbf{0.995$\pm$0.031}      & \textbf{0.801$\pm$0.019}        \\ \hline\bottomrule
\multicolumn{7}{c}{ETTh2}                                                                                                                                                                                         \\ \hline
\multicolumn{1}{c|}{LSTNet}     & 1.536$\pm$0.129          & -          & 1.712$\pm$0.050          & -          & 2.757$\pm$0.089          & 2.002$\pm$0.025          \\
\multicolumn{1}{c|}{IMV-tensor}     & 1.169$\pm$0.076          & -          & 1.874$\pm$0.122          & -          & 2.402$\pm$0.092          & 1.815$\pm$0.085          \\
\multicolumn{1}{c|}{Latent-ODE}     & 3.218$\pm$0.613               & 3.461$\pm$0.568                 & 3.669$\pm$0.510                & 3.900$\pm$0.463                   & 4.077$\pm$0.427               & 3.665$\pm$0.498                 \\
\multicolumn{1}{c|}{Latent ODE-RNN} & 2.911$\pm$0.440                & 3.074$\pm$0.393                 & 3.160$\pm$0.353                & 3.306$\pm$0.315                 & 3.367$\pm$0.292               & 3.164$\pm$0.354                 \\
\multicolumn{1}{c|}{ETN-ODE}        & 1.207$\pm$0.184               & 1.512$\pm$0.141                 & 1.805$\pm$0.142               & 2.117$\pm$0.150                  & 2.410$\pm$0.156                & 1.810$\pm$0.145                  \\
\multicolumn{1}{c|}{\textbf{EgPDE-Net}} & \textbf{0.883$\pm$0.021}      & \textbf{1.222$\pm$0.020}         & \textbf{1.526$\pm$0.018}      & \textbf{1.834$\pm$0.016}        & \textbf{2.078$\pm$0.018}      & \textbf{1.509$\pm$0.014}        \\ \hline\bottomrule
\multicolumn{7}{c}{Electricity}                                                                                                                                                                                   \\ \hline
\multicolumn{1}{c|}{LSTNet}     & 6.058$\pm$0.412          & -          & 4.253$\pm$0.164          & -          & 8.821$\pm$0.416          & 6.378$\pm$0.099          \\
\multicolumn{1}{c|}{IMV-tensor}     & 2.701$\pm$0.255          & -          & 3.719$\pm$0.147          & -          & 4.336$\pm$0.094          & 3.585$\pm$0.141          \\
\multicolumn{1}{c|}{Latent-ODE}     & 10.716$\pm$1.721              & 10.966$\pm$1.885                & 11.312$\pm$2.160               & 11.666$\pm$2.433                & 12.085$\pm$2.750               & 11.349$\pm$2.172                \\
\multicolumn{1}{c|}{Latent ODE-RNN} & 6.042$\pm$1.166               & 6.235$\pm$1.354                 & 6.431$\pm$1.567               & 6.661$\pm$1.818                 & 6.849$\pm$1.985               & 6.443$\pm$1.576                 \\
\multicolumn{1}{c|}{ETN-ODE}        & 2.530$\pm$0.174                & 3.062$\pm$0.111                 & 3.374$\pm$0.110                & 3.763$\pm$0.120                  & 3.949$\pm$0.082               & 3.336$\pm$0.110                  \\
\multicolumn{1}{c|}{\textbf{EgPDE-Net}} & \textbf{2.188$\pm$0.071}      & \textbf{2.835$\pm$0.153}        & \textbf{2.988$\pm$0.449}      & \textbf{3.573$\pm$0.159}        & \textbf{3.72$\pm$0.140}        & \textbf{3.101$\pm$0.135}        \\ \hline\bottomrule
\end{tabular}}
\label{tab-arbi-2}
\end{table*}
\subsection{Baselines}
Three continuous-time deep learning methods and two non-continuous methods are chosen for the comparisons:\\
\textbf{Latent ODE}~\cite{chen2018neural}: It uses a variational autoencoder structure with reversed time inputs. The encoder and decoder networks are both RNNs.\\
\textbf{Latent ODE-RNN}~\cite{rubanova2019latent}: This method adds the ODE net into the RNN internal structure and uses the ODE-RNN framework as the encoder for time series classification and regression. The hidden states are generated with the ODE net before updated in the RNN cell.\\
\textbf{ETN-ODE}~\cite{gao2022explainable}: This method leverages tensorized GRU and tandem attention to encode the raw time series and applies the ODE net to produce multiple future values at arbitrary time points.\\
\textbf{MTGODE}~\cite{jin2022multivariate}: This method first abstracts multivariate time series into dynamic graphs with time-evolving node features and unknown graph structures. Then it leverages neural ODE to process the graph features with continuous encoder.\\
\textbf{STG-NCDE}~\cite{choi2022graph}: This method extends the concept of neural controlled differential equation (NCDE) and designs two NCDEs: one for the temporal processing and the other for the spatial processing. The two NCDEs are combined into a single framework for traffic forecasting.\\
\textbf{STGODE}~\cite{FangLSX21}: This method captures spatial-temporal dynamics through a tensor-based ordinary differential equation and uses a temporal dialated convolution structure to represent long term temporal dependencies for traffic forecasting.\\
\textbf{IMV-tensor}~\cite{guo2019exploring}: It employs a tensorized LSTM to capture different dynamics in multivariate time series and mixture attention to model the generative process of the target series for the next value prediction.\\
\textbf{LSTNet}~\cite{lai2018modeling}: It combines the CNN and RNN to extract short-term local dependency patterns among variables and to discover long-term patterns for time series trends. The aim is to forecast the multi-step future value of each individual variable for multivariate time series. We add one linear layer to output all the future values for multi-step prediction.\\

\subsection{Results of arbitrary-step prediction}
For the arbitrary-step prediction task, we follow the settings of \cite{gao2022explainable}. The basic idea for arbitrary-step prediction is to forecast the multiple future values at arbitrary time steps which are not recorded in the original time series. The output could be arbitrary multiple values between two observations of a fixed sample gap in the test stage. For instance, the electricity consumption data is sampled hourly. Given the historical data, our model could output the future values in the next thirty minutes or the next one and a half hours. We could adjust the integrated time interval to obtain desired future values based on Eq.~\ref{eq10}. In the experimental parts, since there are no public datasets specifically adapted to forecasting arbitrary future values, we re-sample the dataset to half of its original size by taking twice the sampling gap to better and more conveniently demonstrate and verify the effectiveness of arbitrary-step prediction quantitatively. The model only outputs three future values at integral time points sharing the same sample gap as the input data during the training stage, e.g. $T + 1$; $T + 2$; $T + 3$. In the testing stage, the model would output two additional future values at continuous steps, e.g. $T+1.5$ and $T+2.5$, which are not involved during training.

Table~\ref{tab-arbi-1} and Table~\ref{tab-arbi-2} show the RMSE and MAE of arbitrary-step prediction on the four datasets compared with both continuous and non-continuous baseline methods respectively. The tables contain the error of each step with integral time points ``Step1", ``Step2", ``Step3", and continuous-time points ``Step1.5" and ``Step2.5". The column ``Average" represents the mean error of the five steps. For the IMV-tensor and LSTNet methods, we did not report the results of continuous-time points because of its non-continuous model limitation. The results demonstrate that our proposed model EgPDE-Net achieves the best performance among continuous and non-continuous methods. EgPDE-Net obtains the smallest RMSE and MAE on each predicted time step on the four datasets. Latent-ODE and Latent ODE-RNN have relatively large errors on RMSE and MAE, in which the encoder structure has limitations for capturing the relationship among multivariate time series. Our proposed model EgPDE-Net outperforms the best baseline model ETN-ODE by achieving an average decrease of the five-time steps of $16.39\%, 4.24\%, 11.95\%, 5.73\%$ on RMSE and $24.64\%, 8.35\%, 16.63\%, 7.04\%$ on MAE for SML2010, ETTh1, ETTh2 and Electricity datasets respectively.
The designed architecture successfully transforms the PDE problem into the ODE problem which could be solved by the ODE black-box solver tractably. The first ODE net captures the inter-series correlation among the exogenous variables, which is considered as the regularisation term. The second ODE net captures the local information for the target series conditioned on the regularised partial derivatives. Our proposed model EgPDE-Net both utilises the global relative information among different series and local temporal information in each individual series.

\begin{figure}[t]
    \centering
    \includegraphics[width=0.5\textwidth]{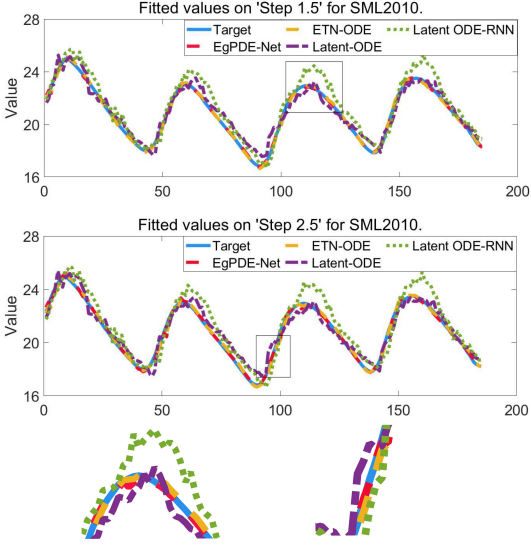}
    \caption{Visualization of arbitrary-step prediction on forecasting ``Step1.5'' and ``Step2.5'' for the SML2010 dataset.}
    \label{smlarbi}
\end{figure}
\begin{figure}[t]
    \centering
    \includegraphics[width=0.5\textwidth]{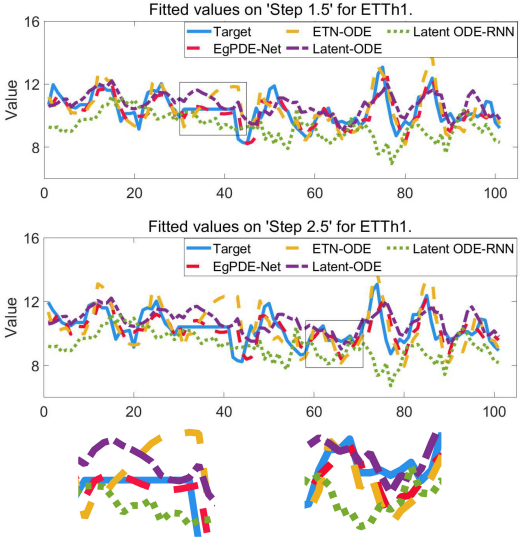}
    \caption{Visualization of arbitrary-step prediction on forecasting ``Step1.5'' and ``Step2.5'' for the ETTh1 dataset.}
    \label{etth1arbi}
\end{figure}
\begin{figure}[t]
    \centering
    \includegraphics[width=0.5\textwidth]{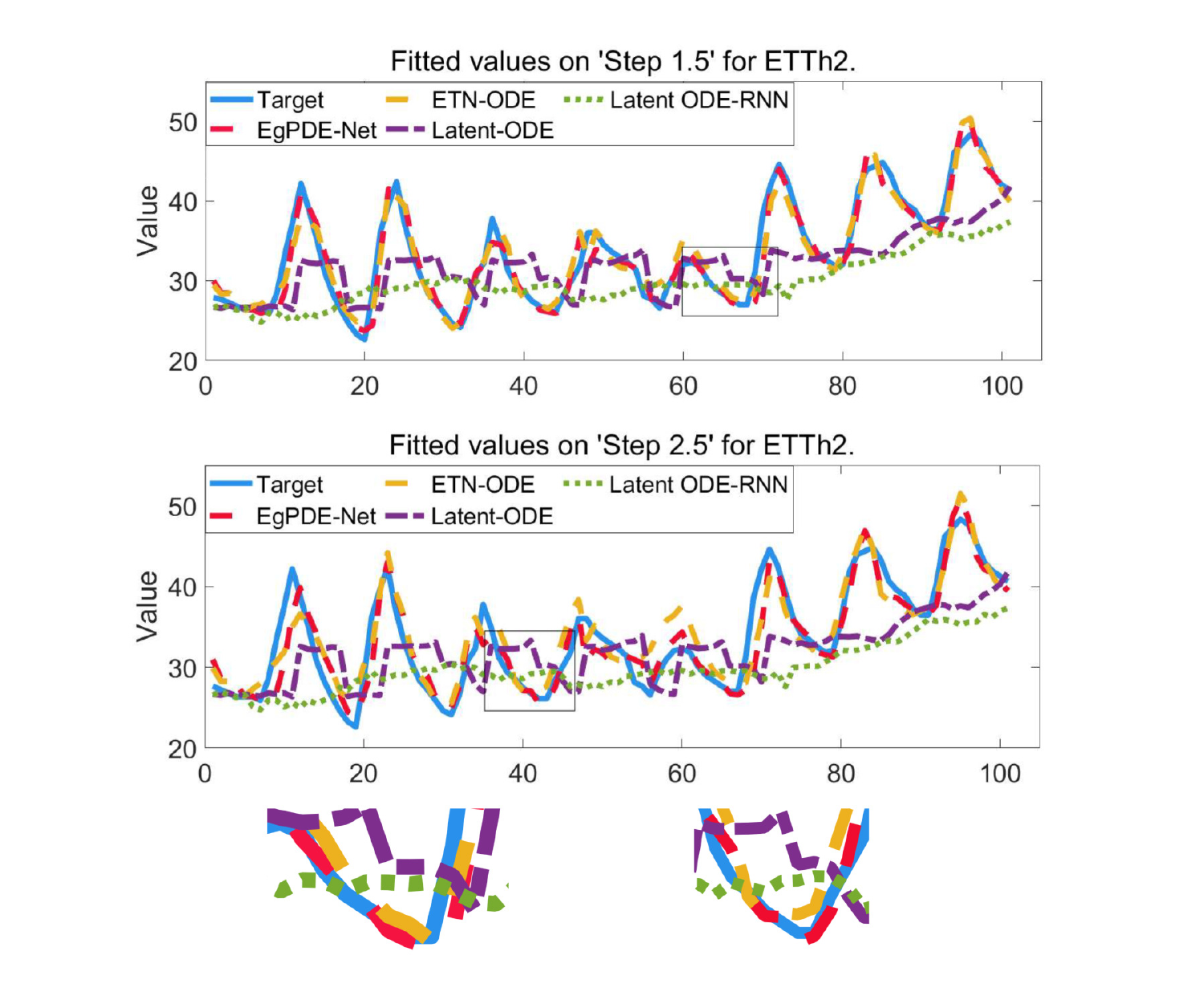}
    \caption{Visualization of arbitrary-step prediction on forecasting ``Step1.5'' and ``Step2.5'' for the ETTh2 dataset.}
    \label{etth2arbi}
\end{figure}
\begin{figure}[t]
    \centering
    \includegraphics[width=0.5\textwidth]{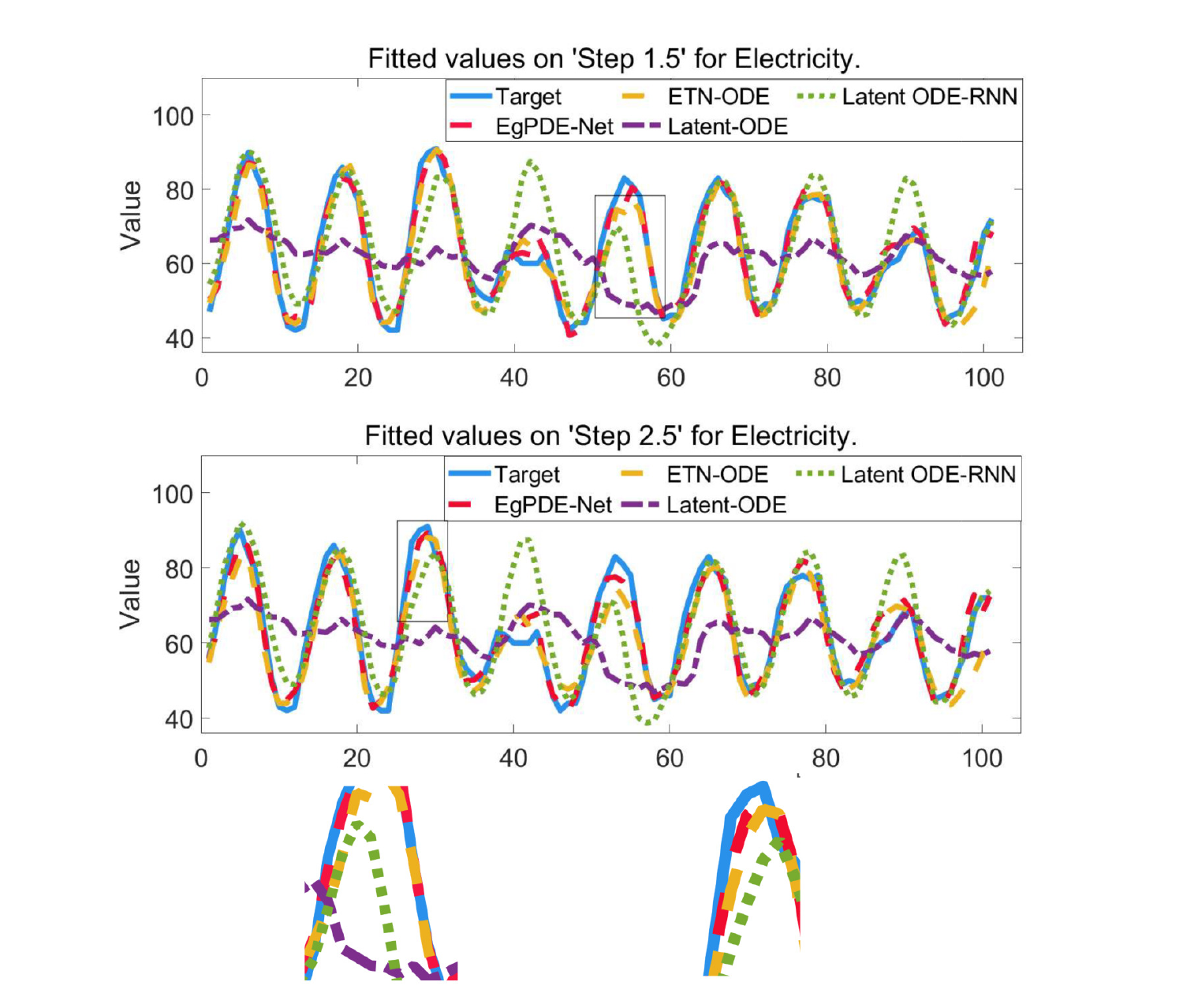}
    \caption{Visualization of arbitrary-step prediction on forecasting ``Step1.5'' and ``Step2.5'' for the Electricity dataset.}
    \label{elecarbi}
\end{figure}

In Fig.~\ref{smlarbi}-\ref{elecarbi}, we visualize the two extra predicted values on-time point $T+1.5$ and $T+2.5$ of the target series on the four datasets. The rectangular regions are enlarged to show the prediction effects of each method more clearly. The red dashed line represents the forecasting values of EgPDE-Net. The original target series of SML2010 and Electricity datasets has a periodic tendency. Fig.~\ref{smlarbi} and Fig.~\ref{elecarbi} show that both EgPDE-Net and ETN-ODE fit the target series perfectly. However, our proposed model EgPDE-Net forecasts the target series better in the crests and troughs on SML2010 and Electricity datasets. In Fig.~\ref{etth1arbi} and Fig.~\ref{etth2arbi}, we could recognise more clearly that the red dashed line is closer to the target series described by the solid blue line, which means the EgPDE-Net model has better fitting results. For example, the red dashed line successfully captures the stable period on time intervals 30 to 40 in Fig.~\ref{etth1arbi}.

\begin{figure}[!htbp]
	\centering
	\subfigure[Results on SML dataset.]{
		\label{vari.sub1}
		\includegraphics[width=0.45\textwidth]{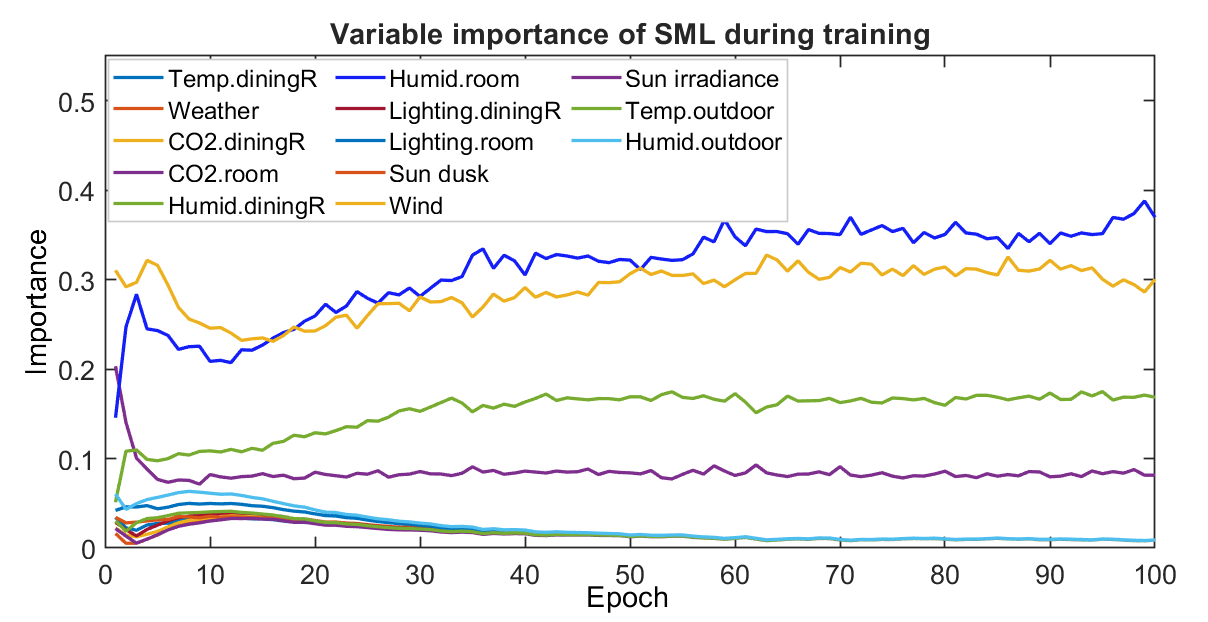}}
	\quad
	\subfigure[Results on Electricity dataset.]{
		\label{vari.sub2}
		\includegraphics[width=0.45\textwidth]{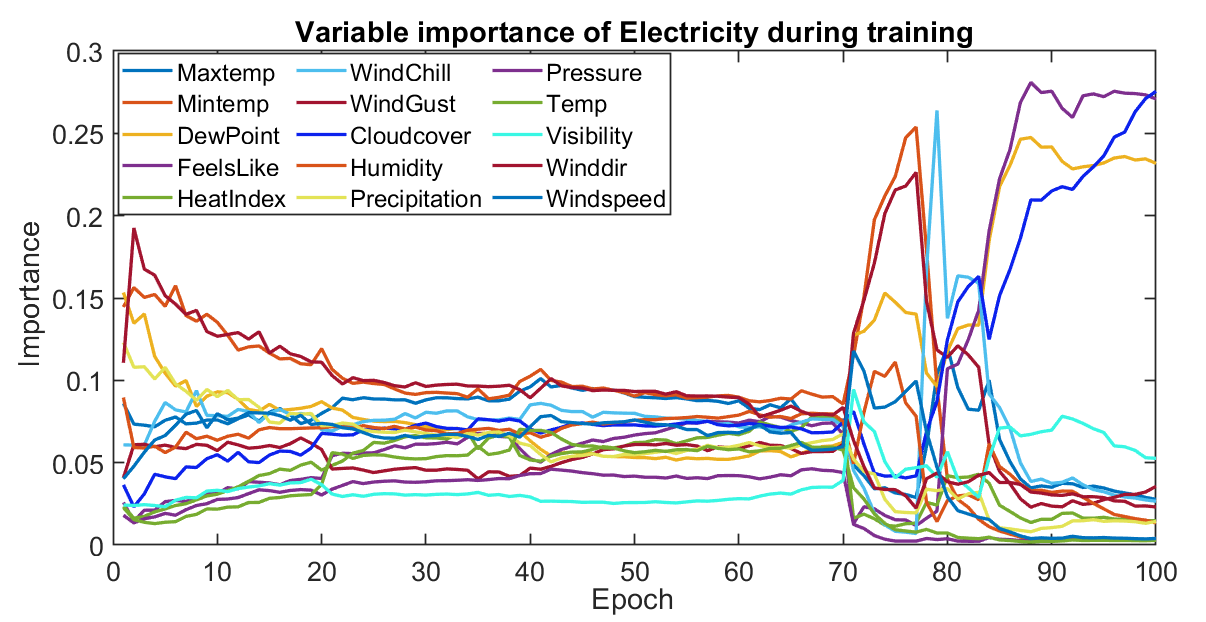}}
\caption{Variable contribution of arbitrary-step prediction.}
	\label{fig:vari}
\end{figure}

{Furthermore, we visualize the variable contribution in Fig.~\ref{fig:vari} on SML and Electricity datasets for arbitrary-step prediction. The overall results show that different variable has a different impact on the target series, which is consistent with our life experience.
For the SML dataset, variables ``Humid.room" and ``CO2.diningR" have larger impacts on the target series during training. In our lives, the amount of carbon dioxide will affect the temperature of the room. More carbon dioxide will make the temperature of the room higher. Humidity affects the efficiency of indoor appliances, such as air conditioners. When the air humidity is high, the air conditioner requires more energy to remove moisture from the air, which affects the temperature change in the room.
For the electricity dataset, variables ``FeelsLike", ``Cloudcover", and ``DewPoint" have larger impacts on the target series ``electricity consumption". In the real world, ``FeelsLike" is a comprehensive consideration of humidity, wind speed, temperature and other factors to describe the actual temperature that people feel. When the perceived temperature is higher than the actual temperature, people may use air conditioning more frequently to lower the temperature, resulting in increased electricity consumption. Conversely, when the perceived temperature is lower than the actual temperature, people may use more heating, which also increases electricity consumption. Cloud cover affects temperature and sunlight exposure, which indirectly affects household electricity consumption. For example, cloud cover reduces sunlight exposure and can lead to the need for more indoor lighting, which increases electricity consumption. The dew point temperature is the temperature at which air reaches saturation and begins to condense at a certain pressure. High dew points may prompt people to use dehumidifiers, while low dew points may prompt people to use humidifiers. Both devices will increase electricity consumption.}

\begin{table*}[t]
\centering
\caption{{Forecasting results of standard multi-step with $M\in\{1, 5, 10\}$ on different datasets.}}
\scalebox{0.9}{
\begin{tabular}{cllllll}
\toprule\hline
                                    & RMSE                 & MAE                  & RMSE                 & MAE                  & RMSE                 & MAE                  \\ \cline{2-7}
                                    & \multicolumn{2}{c}{M=1}                     & \multicolumn{2}{c}{M=5}                     & \multicolumn{2}{c}{M=10}                    \\ \cline{2-7}
\multicolumn{7}{c}{SML2010}                                                                                                                                                   \\ \hline
\multicolumn{1}{c|}{LSTNet}     & \textbf{0.051$\pm$0.012}          & \textbf{0.040$\pm$0.010}          & 0.367$\pm$0.111          & 0.283$\pm$0.089          & 0.563$\pm$0.057           & 0.413$\pm$0.036          \\
\multicolumn{1}{c|}{IMV-tensor}     & 0.122$\pm$0.026          & 0.093$\pm$0.029          & 0.186$\pm$0.049          & 0.134$\pm$0.041          & 0.298$\pm$0.031           & 0.209$\pm$0.017          \\
\multicolumn{1}{c|}{MTGODE}     & 0.120$\pm$0.015          & 0.091$\pm$0.009          & 0.321$\pm$0.034          & 0.215$\pm$0.017          & 0.597$\pm$0.081           & 0.388$\pm$0.031          \\
\multicolumn{1}{c|}{STG-NCDE}     & 0.129$\pm$0.049          & 0.110$\pm$0.046          & 0.192$\pm$0.032          & 0.154$\pm$0.027          & 0.210$\pm$0.034           & 0.157$\pm$0.031          \\
\multicolumn{1}{c|}{STGODE}     & 0.063$\pm$0.013          & 0.050$\pm$0.010          & 0.122$\pm$0.005          & 0.093$\pm$0.006          & 0.216$\pm$0.011           & 0.158$\pm$0.008          \\
\multicolumn{1}{c|}{Latent-ODE}     & 0.591$\pm$0.062          & 0.445$\pm$0.036          & 0.674$\pm$0.075          & 0.505$\pm$0.054          & 0.860$\pm$0.112           & 0.643$\pm$0.082          \\
\multicolumn{1}{c|}{Latent ODE-RNN} & 0.469$\pm$0.100            & 0.378$\pm$0.077          & 0.513$\pm$0.119          & 0.415$\pm$0.098          & 0.557$\pm$0.075          & 0.438$\pm$0.075          \\
\multicolumn{1}{c|}{ETN-ODE}        & 0.066$\pm$0.018          & 0.052$\pm$0.017          & 0.127$\pm$0.012          & 0.091$\pm$0.006          & 0.234$\pm$0.029          & 0.161$\pm$0.018          \\
\multicolumn{1}{c|}{\textbf{EgPDE-Net}} & 0.058$\pm$0.018 & 0.044$\pm$0.014 & \textbf{0.117$\pm$0.005} & \textbf{0.085$\pm$0.004} & \textbf{0.195$\pm$0.006} & \textbf{0.138$\pm$0.003} \\ \hline\bottomrule
\multicolumn{7}{c}{ETTh1}                                                                                                                                                     \\ \hline
\multicolumn{1}{c|}{LSTNet}     & 0.640$\pm$0.009          & 0.427$\pm$0.007          & 1.077$\pm$0.018          & 0.755$\pm$0.019          & 1.454$\pm$0.058           & 1.079$\pm$0.069          \\
\multicolumn{1}{c|}{IMV-tensor}        & 0.658$\pm$0.037          & 0.459$\pm$0.035          & 1.097$\pm$0.021          & 0.777$\pm$0.027           & 1.424$\pm$0.033          & 1.032$\pm$0.051          \\
\multicolumn{1}{c|}{MTGODE}     & 0.627$\pm$0.006          & 0.424$\pm$0.008          & 1.018$\pm$0.007          & 0.696$\pm$0.010          & 1.323$\pm$0.021           & 0.934$\pm$0.026          \\
\multicolumn{1}{c|}{STG-NCDE}     & 0.642$\pm$0.013          & 0.425$\pm$0.010          & 1.021$\pm$0.015          & 0.705$\pm$0.024          & 1.331$\pm$0.026           & 0.936$\pm$0.024          \\
\multicolumn{1}{c|}{STGODE}     & 0.639$\pm$0.011          & 0.463$\pm$0.015          & 1.070$\pm$0.039          & 0.780$\pm$0.031          & 1.434$\pm$0.090           & 1.076$\pm$0.064          \\
\multicolumn{1}{c|}{Latent-ODE}     & 0.928$\pm$0.035          & 0.683$\pm$0.028          & 1.193$\pm$0.106          & 0.870$\pm$0.102           & 1.751$\pm$0.238          & 1.322$\pm$0.206          \\
\multicolumn{1}{c|}{Latent ODE-RNN} & 1.592$\pm$0.232          & 1.252$\pm$0.226          & 1.243$\pm$0.091          & 0.951$\pm$0.083          & 2.029$\pm$0.224          & 1.656$\pm$0.222          \\
\multicolumn{1}{c|}{ETN-ODE}        & 0.635$\pm$0.007          & 0.434$\pm$0.009          & 1.049$\pm$0.027          & 0.748$\pm$0.030           & 1.394$\pm$0.059          & 1.015$\pm$0.053          \\
\multicolumn{1}{c|}{\textbf{EgPDE-Net}} & \textbf{0.627$\pm$0.003} & \textbf{0.414$\pm$0.004} & \textbf{1.013$\pm$0.016} & \textbf{0.685$\pm$0.017} & \textbf{1.309$\pm$0.012} & \textbf{0.914$\pm$0.009} \\ \hline\bottomrule
\multicolumn{7}{c}{ETTh2}                                                                                                                                                     \\ \hline
\multicolumn{1}{c|}{LSTNet}     & 0.732$\pm$0.013          & 0.499$\pm$0.011          & 2.480$\pm$0.070          & 1.676$\pm$0.039          & 3.939$\pm$0.076           & 2.849$\pm$0.052          \\
\multicolumn{1}{c|}{IMV-tensor}        & 0.831$\pm$0.047          & 0.605$\pm$0.044          & 2.184$\pm$0.114           & 1.502$\pm$0.125          & 3.202$\pm$0.074          & 2.256$\pm$0.055          \\
\multicolumn{1}{c|}{MTGODE}     & 1.004$\pm$0.095          & 0.710$\pm$0.061          & 2.545$\pm$0.029          & 1.681$\pm$0.029          & 3.607$\pm$0.018           & 2.468$\pm$0.019          \\
\multicolumn{1}{c|}{STG-NCDE}     & 0.810$\pm$0.141          & 0.574$\pm$0.133          & 2.021$\pm$0.086          & 1.320$\pm$0.047          & 2.980$\pm$0.124           & 1.986$\pm$0.094          \\
\multicolumn{1}{c|}{STGODE}     & 0.711$\pm$0.016          & 0.510$\pm$0.021          & 1.876$\pm$0.030          & 1.264$\pm$0.031          & 2.870$\pm$0.015           & 1.990$\pm$0.054          \\
\multicolumn{1}{c|}{Latent-ODE}     & 2.208$\pm$0.336          & 1.638$\pm$0.267          & 3.667$\pm$0.747          & 2.945$\pm$0.885          & 4.110$\pm$0.473           & 3.025$\pm$0.461          \\
\multicolumn{1}{c|}{Latent ODE-RNN} & 4.198$\pm$1.317          & 3.399$\pm$1.082          & 4.118$\pm$0.719          & 3.269$\pm$0.742          & 4.229$\pm$0.298          & 3.239$\pm$0.325          \\
\multicolumn{1}{c|}{ETN-ODE}        & 0.732$\pm$0.009          & 0.497$\pm$0.008          & 2.038$\pm$0.040           & 1.357$\pm$0.035          & 3.179$\pm$0.023          & 2.209$\pm$0.055          \\
\multicolumn{1}{c|}{\textbf{EgPDE-Net}} & \textbf{0.719$\pm$0.006} & \textbf{0.496$\pm$0.008} & \textbf{1.871$\pm$0.036} & \textbf{1.193$\pm$0.028} & \textbf{2.863$\pm$0.037} & \textbf{1.889$\pm$0.028} \\ \hline\bottomrule
\multicolumn{7}{c}{Electricity}                                                                                                                                               \\ \hline
\multicolumn{1}{c|}{LSTNet}     & 1.926$\pm$0.015          & \textbf{1.310$\pm$0.019}          & 7.252$\pm$0.183          & 5.480$\pm$0.130          & 11.989$\pm$0.240           & 9.426$\pm$0.162          \\
\multicolumn{1}{c|}{IMV-tensor}        & 2.197$\pm$0.165          & 1.568$\pm$0.166          & 4.371$\pm$0.267          & 3.166$\pm$0.177           & 5.458$\pm$0.216          & 3.963$\pm$0.162          \\
\multicolumn{1}{c|}{MTGODE}     & 2.449$\pm$0.029          & 1.772$\pm$0.034          & 6.096$\pm$0.167          & 4.301$\pm$0.113          & 6.888$\pm$0.307           & 5.094$\pm$0.226          \\
\multicolumn{1}{c|}{STG-NCDE}     & 2.605$\pm$0.252          & 1.967$\pm$0.254          & 4.703$\pm$0.331          & 3.332$\pm$0.201          & 5.405$\pm$0.631           & 3.867$\pm$0.444          \\
\multicolumn{1}{c|}{STGODE}     & 1.930$\pm$0.038          & 1.387$\pm$0.048          & 4.036$\pm$0.091          & 2.931$\pm$0.080          & 5.230$\pm$0.068           & 3.868$\pm$0.080          \\
\multicolumn{1}{c|}{Latent-ODE}     & 8.490$\pm$1.516           & 6.726$\pm$1.271          & 10.025$\pm$2.848         & 8.042$\pm$2.555          & 10.509$\pm$1.96          & 8.299$\pm$1.732          \\
\multicolumn{1}{c|}{Latent ODE-RNN} & 4.308$\pm$0.508          & 3.374$\pm$0.434          & 4.982$\pm$0.735          & 3.909$\pm$0.615          & 5.411$\pm$0.214          & 4.265$\pm$0.194          \\
\multicolumn{1}{c|}{ETN-ODE}        & 2.277$\pm$0.045          & 1.684$\pm$0.045          & 4.094$\pm$0.116          & 2.873$\pm$0.030           & 5.084$\pm$0.234          & 3.598$\pm$0.139          \\
\multicolumn{1}{c|}{\textbf{EgPDE-Net}} & \textbf{1.923$\pm$0.028} & 1.346$\pm$0.033 & \textbf{3.672$\pm$0.199} & \textbf{2.570$\pm$0.164}  & \textbf{4.668$\pm$0.128} & \textbf{3.372$\pm$0.107} \\ \hline\bottomrule
\end{tabular}}
\label{tab-multi}
\end{table*}

\subsection{Results of standard multi-step prediction}
In this section, we compare the performance of various methods on three different standard multi-step prediction tasks, forecasting the next 1, 5 and 10 future values testifying the ability of predicting short and long term sequence. Table~\ref{tab-multi} shows the RMSE and MAE of all the methods on the four datasets. The best results are displayed boldfaced.
In most cases, we could observe that the proposed model EgPDE-Net achieves the smallest errors on both RMSE and MAE metrics. It indicates the success of two ODE nets modelling the inter-series correlation among exogenous variables and intra-series relationship in the temporal aspect. Specifically, EgPDE-Net has better performance on long-term forecasting tasks than other methods. This result might be contributed to the representation of the correlation of exogenous variables in EgPDE-Net, which could influence the prediction of the target series in the long-term period. Focusing on predicting one target series and modelling the influence of other exogenous variables could improve the performance of multi-step prediction.
{Although LSTNet achieves smaller RMSE and MAE, our method also produces competitive results on task $M=1$ in the SML2010 dataset. LSTNet aims to forecast the desirable future value ahead of the current time stamp at a specific time point. This model outputs one value at a time, which has an advantage in predicting one-step future value by capturing the local information with CNN structure. It focuses on one-step short-term forecasting without considering the continuous changes among the multiple future values. For long-term forecasting tasks, LSTNet has unsatisfactory performance. This model outputs multiple future values with a linear layer and has limitations in processing accumulative errors without modeling the forecasting incremental information. Our EgPDE-Net has the advantage of forecasting one target series with the input of multivariate time series other than outputting each individual variable.}
Neither Latent ODE nor Latent ODE-RNN are performing well in multi-step prediction, both of which focus more on the reconstruction of time series and temporal relationships. The RNN encoder in Latent ODE brings limitations in processing multivariate time series. The Latent ODE-RNN model adds the ODE net in the RNN basic cell and has advantages in dealing with irregular sampling data. However, this framework performs worse when the predicted target series is affected by other complex exogenous variables.
{Compared with these graph-based continuous methods equipped with neural ODE, our method achieves promising results and reflects the success of capturing interacting information among exogenous variables with the PDE framework. These graph-based methods all leverage neural ODE as an encoder, and the decoder is normal linear networks, which has limitations for multi-step prediction and has unsatisfactory performance in modeling the incremental change with small cumulative errors.}
Compared with the non-continuous method IMV-tensor, our proposed method keeps the advantage of predicting short and long term period future values. These results indicate that building continuous networks utilising the information of exogenous variables could benefit arbitrary-step prediction and improve model performance for standard multi-step prediction. {We further conduct extra experiments on the forecasting step M=20 to testify the performance of long-term forecasting with several competitive models. The results are shown in Table~\ref{M20}. We can see that our method is also effective and achieves the best performance on most metrics of the four datasets for long-term forecasting.}
\begin{table*}[t]
\centering
\caption{Compared results on SML, Electricity, ETTh1, and ETTh2 datasets with forecasting step M=20.}
\begin{tabular}{c|cc|cc|cc|cc}
\toprule
Dataset   & \multicolumn{2}{c|}{SML} & \multicolumn{2}{c|}{Electricity} & \multicolumn{2}{c|}{ETTh1} & \multicolumn{2}{c}{ETTh2} \\ \hline
Metrics   & RMSE        & MAE        & RMSE            & MAE            & RMSE         & MAE         & RMSE        & MAE         \\ \hline
STGODE    & 0.511$\pm$0.031  & 0.364$\pm$0.029 & 6.069$\pm$0.085      & 4.587$\pm$0.061     & 1.675$\pm$0.035   & 1.304$\pm$0.025  & \textbf{3.612$\pm$0.027}  & 2.647$\pm$0.023  \\
MTGODE    & 1.146$\pm$0.040  & 0.843$\pm$0.058 & 6.014$\pm$0.164      & 4.423$\pm$0.110     & 1.605$\pm$0.018   & 1.181$\pm$0.022  & 3.779$\pm$0.024  & 2.607$\pm$0.030  \\
STG-NCDE  & 0.423$\pm$0.023  & 0.299$\pm$0.024 & 5.613$\pm$0.217      & 4.128$\pm$0.151     & 1.690$\pm$0.029   & 1.250$\pm$0.056  & 3.697$\pm$0.065  & 2.646$\pm$0.065  \\
ETN-ODE   & 0.508$\pm$0.060  & 0.369$\pm$0.054 & 5.715$\pm$0.085      & 4.239$\pm$0.064     & 1.705$\pm$0.055   & 1.279$\pm$0.049  & 3.958$\pm$0.170  & 2.921$\pm$0.122  \\
EgPDE-Net & \textbf{0.396$\pm$0.038}  & \textbf{0.261$\pm$0.017} & \textbf{5.493$\pm$0.126}      & \textbf{3.999$\pm$0.079}     & \textbf{1.600$\pm$0.019}   & \textbf{1.177$\pm$0.022}  & 3.648$\pm$0.023  & \textbf{2.582$\pm$0.024}  \\ \bottomrule
\end{tabular}\label{M20}
\end{table*}

\subsection{Ablation study}
In this section, we design two variants of EgPDE-Net to demonstrate the effectiveness and importance of our model components.
\begin{itemize}
    \item \textbf{w/o self-att:} Replace the self attention component with GRU layer when embedding the exogenous variables.
    \item \textbf{w/o zx\_ode:} Remove the first ODE net and use an LSTM layer to obtain the weight for each forecasting step.
\end{itemize}
We conduct the ablation experiments on the arbitrary-step prediction task on the four datasets, and the results of RMSE is shown in Table~\ref{tab-ablation}. In general, the complete EgPDE-Net achieves the best performance on all the datasets. Removing any component of EgPDE-Net will increase the forecasting errors. Analysing the mean errors of the five steps on the column ``Average'', the forecasting error increases much without the first ODE net, which aims to deal with the exogenous variables. This result indicates that it is effective to process target series and exogenous variables separately and merge the information in latent space. Replacing self-attention also leads to increased forecasting errors. The attention mechanism could allocate different contributions among the exogenous variables and generate a more adaptive input for the first ODE net.

\begin{table}[h]
\centering
\caption{RMSE of variants of the EgPDE-Net model on the arbitrary-step prediction task.}
\scalebox{0.65}{
\begin{tabular}{cllllll}
\toprule\hline
                                    & \multicolumn{1}{c}{Step1} & \multicolumn{1}{c}{Step1.5} & \multicolumn{1}{c}{Step2} & \multicolumn{1}{c}{Step2.5} & \multicolumn{1}{c}{Step3} & \multicolumn{1}{c}{Average} \\ \cline{2-7}
\multicolumn{7}{c}{SML2010}                                                                                                                                                                                       \\ \hline
\multicolumn{1}{c|}{w/o self-att}   & 0.121$\pm$0.011               & 0.156$\pm$0.011                 & 0.184$\pm$0.012               & 0.212$\pm$0.013                 & 0.239$\pm$0.016               & 0.187$\pm$0.012                 \\
\multicolumn{1}{c|}{w/o zx\_ode}    & 0.107$\pm$0.024               & 0.125$\pm$0.022                 & 0.149$\pm$0.015               & 0.195$\pm$0.024                 & 0.273$\pm$0.044               & 0.181$\pm$0.016                 \\
\multicolumn{1}{c|}{\textbf{EgPDE-Net}} & \textbf{0.099$\pm$0.007}      & \textbf{0.120$\pm$0.007}        & \textbf{0.145$\pm$0.008}      & \textbf{0.171$\pm$0.010}        & \textbf{0.200$\pm$0.012}       & \textbf{0.151$\pm$0.008}        \\ \hline\bottomrule
\multicolumn{7}{c}{ETTh1}                                                                                                                                                                                         \\ \hline
\multicolumn{1}{c|}{w/o self-att}   & 0.891$\pm$0.007               & 1.043$\pm$0.010                  & 1.175$\pm$0.014               & 1.257$\pm$0.023                 & 1.385$\pm$0.035               & 1.163$\pm$0.018                 \\
\multicolumn{1}{c|}{w/o zx\_ode}    & 0.889$\pm$0.020                & 1.058$\pm$0.021                 & 1.210$\pm$0.035                & 1.309$\pm$0.021                 & 1.422$\pm$0.015               & 1.192$\pm$0.016                 \\
\multicolumn{1}{c|}{\textbf{EgPDE-Net}} & \textbf{0.863$\pm$0.009}      & \textbf{1.028$\pm$0.012}        & \textbf{1.162$\pm$0.011}      & \textbf{1.263$\pm$0.015}        & \textbf{1.372$\pm$0.020}       & \textbf{1.151$\pm$0.013}        \\ \hline\bottomrule
\multicolumn{7}{c}{ETTh2}                                                                                                                                                                                         \\ \hline
\multicolumn{1}{c|}{w/o self-att}   & 1.298$\pm$0.034               & 1.837$\pm$0.055                 & 2.310$\pm$0.069                & 2.804$\pm$0.088                 & 3.141$\pm$0.095               & 2.372$\pm$0.070                  \\
\multicolumn{1}{c|}{w/o zx\_ode}    & 1.327$\pm$0.038               & 1.859$\pm$0.010                  & 2.500$\pm$0.111                 & 2.875$\pm$0.095                 & 3.287$\pm$0.175               & 2.472$\pm$0.091                 \\
\multicolumn{1}{c|}{\textbf{EgPDE-Net}} & \textbf{1.272$\pm$0.017}      & \textbf{1.771$\pm$0.015}        & \textbf{2.222$\pm$0.022}      & \textbf{2.680$\pm$0.029}         & \textbf{3.005$\pm$0.039}      & \textbf{2.276$\pm$0.023}        \\ \hline\bottomrule
\multicolumn{7}{c}{Electricity}                                                                                                                                                                                   \\ \hline
\multicolumn{1}{c|}{w/o self-att}   & 3.213$\pm$0.087               & 4.007$\pm$0.092                 & 4.496$\pm$0.084               & 5.046$\pm$0.108                 & 5.392$\pm$0.115               & 4.497$\pm$0.079                 \\
\multicolumn{1}{c|}{w/o zx\_ode}    & 3.073$\pm$0.070                & 3.985$\pm$0.166                 & 4.918$\pm$0.253               & 5.388$\pm$0.287                 & 5.935$\pm$0.363               & 4.772$\pm$0.200                   \\
\multicolumn{1}{c|}{\textbf{EgPDE-Net}} & \textbf{3.036$\pm$0.074}      & \textbf{3.862$\pm$0.167}        & \textbf{4.358$\pm$0.176}      & \textbf{4.819$\pm$0.178}        & \textbf{5.084$\pm$0.143}      & \textbf{4.294$\pm$0.145}        \\ \hline\bottomrule
\end{tabular}}
\label{tab-ablation}
\end{table}

{
\subsection{Discussion and Limitation}
For continuous modeling, it could cost more time in the training stage and inference. We will investigate more into the internal structure of ODE net to accelerate training. In the arbitrary-step prediction, the impact of different resampling sizes on prediction performance could be further investigated. Moreover, in the design of the encoding network, we consider the recurrent neural network which is naturally adaptive to the time series. The encoding network is built as discrete models, which may limit the feature representation with different data types. In the future, we will explore building continuous encoding networks to deal with richer data types.}

\section{Conclusion}
In this work, we proposed the Exogenous-guided Partial Differential Equation Network (EgPDE-Net), which aims to solve the PDE modelling problem in multivariate time series analysis. We developed a neural network to estimate the partial derivative and considered it as a regularised term to guide the generation trajectory of the specific target series. The two ODE networks applied in this framework take advantage of capturing the intra-series temporal patterns and the inter-series correlations jointly among the target series and the exogenous variables. Focusing on the specific target series prediction with multivariate input could take advantage of the influence of the exogenous variables. Experiments on four real-world datasets demonstrated improvements over the baseline methods.
Theoretical exploration on the precision of weak solutions for PDEs is usually difficult and remains as an open problem in the literature. We will also leave such theoretical investigation of the ODE-based weak solutions as future work. {Moreover, dealing with informative missingness and partial observations is also one important research topic and we will consider to explore this systematically in the future.}

\ifCLASSOPTIONcaptionsoff
  \newpage
\fi


\bibliographystyle{IEEEtran}
\bibliography{ref}
\end{document}